\documentclass[10pt,twocolumn,letterpaper]{article}
\usepackage{cvpr}
\usepackage{times}
\usepackage{epsfig}
\usepackage{graphicx}
\graphicspath{ {./images/} }
\usepackage{amsmath}
\usepackage{amssymb}
\usepackage{makecell}
\usepackage{multirow}
\usepackage{color, colortbl}
\usepackage{soul}
\usepackage{stackengine}
\usepackage{csquotes}
\usepackage{amsmath,bm, bbm}
\usepackage[ruled, lined, linesnumbered, longend]{algorithm2e}
\let\oldnl\nl
\newcommand{\nonl}{\renewcommand{\nl}{\let\nl\oldnl}}
\usepackage{amsmath}
\usepackage{subfigure}
\usepackage{caption}
\usepackage{footmisc}
\usepackage[normalem]{ulem}
\usepackage{booktabs,array,siunitx}
\usepackage[table]{xcolor}
\definecolor{Gray}{gray}{0.95}
\newcommand\xrowht[2][0]{\addstackgap[.5\dimexpr#2\relax]{\vphantom{#1}}}

\def\ourencoder{Bilateral Context Module}
\def\ourblock{Bilateral Context Block}
\def\ourblocks{Bilateral Context Blocks}
\def\ourdecoder{Adaptive Fusion Module}
\def\ourextractor{Feature Extractor}

\usepackage[pagebackref=true,breaklinks=true,letterpaper=true,colorlinks,bookmarks=false]{hyperref}

\cvprfinalcopy 

\def\cvprPaperID{1506} 

\begin{document}

\title{Semantic Segmentation for Real Point Cloud Scenes \\via Bilateral Augmentation and Adaptive Fusion}

\author{Shi Qiu$^{1,2}$, Saeed Anwar$^{1,2}$ and Nick Barnes$^{1}$\\
$^1$Australian National University, $^2$Data61-CSIRO, Australia\\
{\tt\small \{shi.qiu, saeed.anwar, nick.barnes\}@anu.edu.au}
}

\maketitle

\begin{abstract}
Given the prominence of current 3D sensors, a fine-grained analysis on the basic point cloud data is worthy of further investigation. Particularly, real point cloud scenes can intuitively capture complex surroundings in the real world, but due to 3D data's raw nature, it is very challenging for machine perception. In this work, we concentrate on the essential visual task, semantic segmentation, for large-scale point cloud data collected in reality. On the one hand, to reduce the ambiguity in nearby points, we augment their local context by fully utilizing both geometric and semantic features in a bilateral structure. On the other hand, we comprehensively interpret the distinctness of the points from multiple resolutions and represent the feature map following an adaptive fusion method at point-level for accurate semantic segmentation. Further, we provide specific ablation studies and intuitive visualizations to validate our key modules. By comparing with state-of-the-art networks on three different benchmarks, we demonstrate the effectiveness of our network.
\end{abstract}

\section{Introduction}

As 3D data acquisition techniques develop rapidly, different types of 3D scanners, \eg LiDAR scanners~\cite{jaboyedoff2012use} and RGB-D cameras~\cite{endres20133} are becoming popular in our daily life. Basically, 3D scanners can capture data that enables AI-driven machines to better see and recognize the world. As a fundamental data representation, point clouds can be easily collected using 3D scanners, retaining abundant information for further investigation. Therefore, point cloud analysis is playing an essential role in 3D computer vision. 

Research has shown great success in terms of basic classification of small-scale point clouds (\ie, objects containing a few thousand points): for example, face ID~\cite{hamza2008face} is now a widely used bio-identification for mobile devices. Researchers have recently been investigating a fine-grained analysis of large and complex point clouds~\cite{thomas2019kpconv, landrieu2018large, hu2020randla, xie2020pointcontrast} because of the tremendous potential in applications such as autonomous driving, augmented reality, robotics, \etc. This paper focuses on the semantic segmentation task to identify each point's semantic label for real point cloud scenes.

\begin{figure}
\begin{center}
\includegraphics[width=0.98\columnwidth]{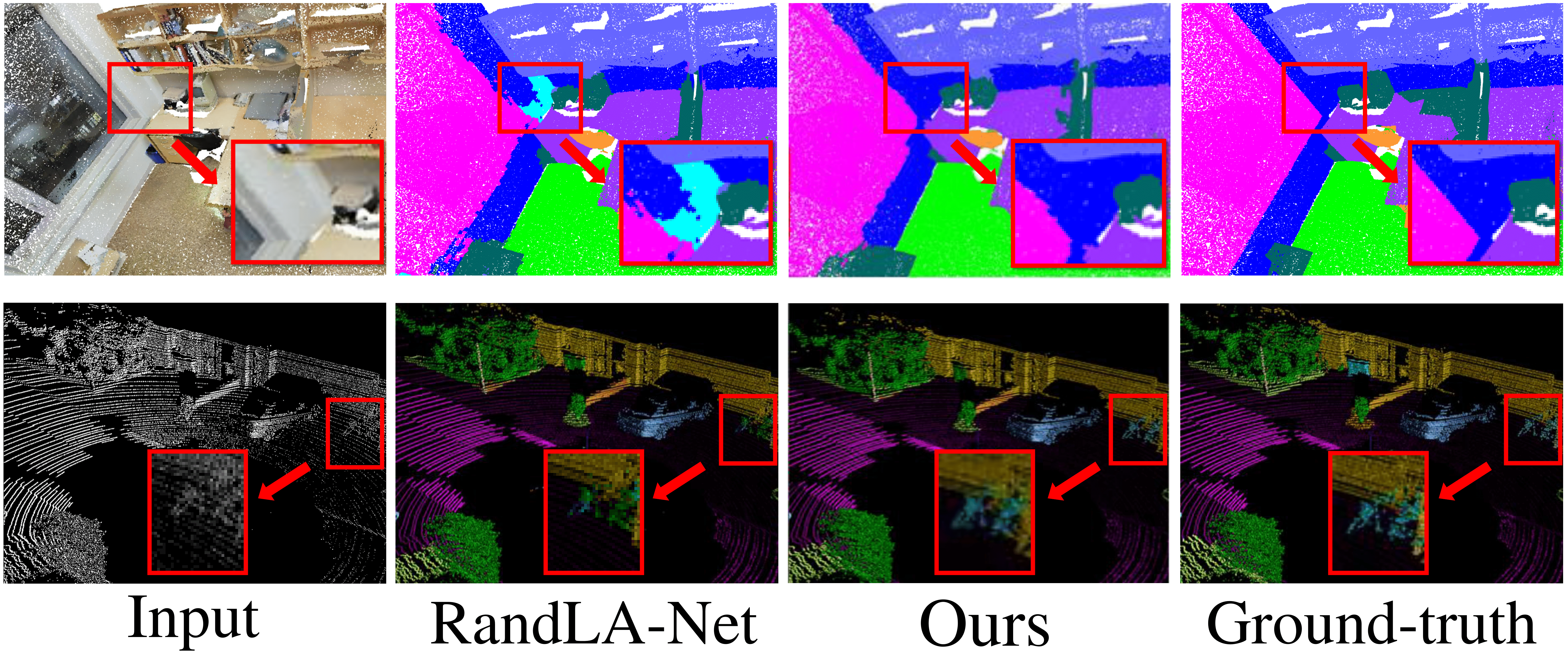}
\end{center}
\vspace{-5mm}
\captionsetup{font=small}
\caption{Examples of semantic segmentation for real point cloud scenes, where the main differences are highlighted and zoomed-in. The upper row shows an \emph{indoor} working environment with $\sim$0.9 million points: RandLA-Net~\cite{hu2020randla} falsely classifies the wall around the room corner, while our result is much closer to the ground-truth. The lower row is an \emph{outdoor} traffic scene containing $\sim$32 thousand points, where a small bike on the right is correctly identified by our network (in blue), while RandLA-Net mislabels it as vegetation (in green).}
\label{fig:compare}
\vspace{-5mm}
\end{figure}
Although there are many notable works~\cite{ronneberger2015u, long2015fully, zhao2017pyramid} addressing the semantic segmentation of 2D images which have a simpler structure, point clouds are \emph{scattered, irregular, unordered, and unevenly distributed} in 3D space, making the corresponding task much more challenging, especially for large scenes made of millions or even billions of points collected from the real world. To deal with the 3D data, many papers try to build data-driven models using deep learning. Specifically, Guo~\etal~\cite{guo2020deep} summarizes the Convolutional Neural Network (CNN) models targeting point clouds into three streams: projection-based, discretization-based, and point-based methods. As a projection-based example, Lawin~\etal~\cite{lawin2017deep} virtually projects 3D point clouds onto images and applies a conventional FCN~\cite{long2015fully} to analyze the 2D multi-view representations. Similarly, the discretization-based approaches model point clouds as voxels~\cite{huang2016point} or lattices~\cite{su2018splatnet} for CNN processing, and finally interpolate the semantic results back to the original input. However, the mentioned methods are not optimal for real applications due to some common issues: firstly, they require several time-consuming pre/post-processing steps to make predictions; and secondly, the generated intermediate representations may partially lose the context of the surroundings. 

To avoid the above issues, in this paper, we prefer point-based networks (details in Sec.~\ref{sec:work}) that directly process the points for fine-grained analysis. Moreover, for an accurate semantic segmentation on real point cloud scenes, we endeavor to resolve the major drawbacks of existing works: 

\emph{Ambiguity in close points.} Most current solutions~\cite{wang2019dynamic, engelmann2020dilated, qiu2021dense} represent a point based on its pre-defined neighbors via a fixed metric like Euclidean distance. However, outliers and overlap between neighborhoods during the neighborhood's construction are difficult to avoid, especially when the points are closely distributed near the boundaries of different semantic classes. To alleviate possible impacts, we attempt to augment the local context by involving a dense region. Moreover, we introduce a robust aggregation process to refine the augmented local context and extract useful neighboring information for the point's representation.

\emph{Redundant features.} We notice an increasing number of works~\cite{hu2020randla, yan2020pointasnl, qiu2019geometric} combine similar features multiple times to enhance the perception of the model. In fact, this process causes redundancy and increases the complexity for the model to process large-scale point clouds. To avoid the above problems, we propose to characterize the input information as geometric and semantic clues and then fully utilize them through a bilateral structure. More compactly, our design can explicitly represent complex point clouds.

\emph{Inadequate global representations.} Although some approaches~\cite{qi2017pointnet++, liu2019relation, li2019pu} apply an encoder-decoder~\cite{badrinarayanan2017segnet} structure to learn the sampled point cloud; the output feature map is inadequate for a fine-grained semantic segmentation analysis since the global perception of the original data would be damaged during the sampling process. In our method, we intend to rebuild such perception by integrating information from different resolutions. Moreover, we adaptively fuse multi-resolutional features for each point to obtain a comprehensive representation, which can be directly applied for semantic prediction.  

To conclude, our contributions are in these aspects:
\begin{itemize}
\item We introduce a bilateral block to augment the local context of the points.
\item We adaptively fuse multi-resolutional features to acquire comprehensive knowledge about point clouds.
\item We present a novel semantic segmentation network using our proposed structures to deal with real point cloud scenes.
\item We evaluate our network on three large-scale benchmarks of real point cloud scenes. The experimental results demonstrate that our approach achieves competitive performances against state-of-the-art methods.
\end{itemize}

\section{Related Work}
\label{sec:work}
\noindent \textbf{Point-Based Approaches:}
As mentioned before, point-based approaches are designed to process unstructured 3D point cloud data directly rather than using its intermediate variants. Particularly, PointNet~\cite{qi2017pointnet} applied the multi-layer-perceptron (MLP) and symmetric function (\eg, max-pooling) to learn and aggregate point cloud features, respectively. Subsequently, point-wise MLPs were used to extract local features based on neighbor searching methods: \eg, ball-query in PointNet++~\cite{qi2017pointnet++}, or k-nearest neighbors (knn) in DGCNN~\cite{wang2019dynamic}. Moreover, MLPs were extended to perform point-convolutions: for instance, KPConv~\cite{thomas2019kpconv} leveraged kernel-points to convolve local point sets, while DPC~\cite{engelmann2020dilated} defined dilated point groups to increase the receptive fields of the points. Recurrent Neural Network (RNN) and Graph Convolutional Network (GCN) have also been adopted to replace regular CNNs in point-based approaches: for example, Liu~\etal~\cite{liu2019point2sequence} transformed point clouds into sequences and processed the scaled areas using an LSTM structure, and Landrieu~\etal~\cite{landrieu2018large} exploited super-point graphs to acquire semantic knowledge.

\begin{figure*}
\begin{center}
\includegraphics[width=0.95\textwidth]{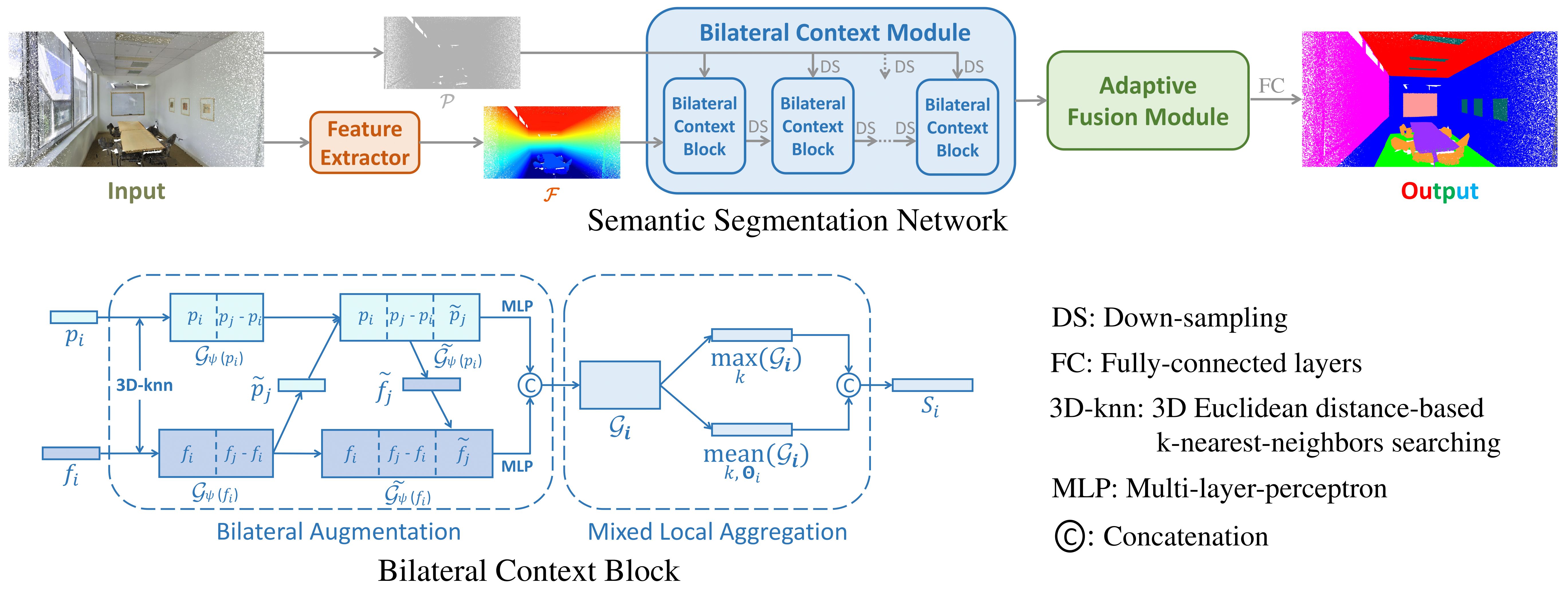}
\end{center}
\vspace{-5mm}
\captionsetup{font=small}
\caption{The details of our semantic segmentation network and the Bilateral Context Block (the annotations are consistent with the items in Sec.~\ref{sec:metho_cross}). Firstly, the Feature Extractor (Sec.~\ref{sec:impl_extractor}) captures the preliminary semantic context $\mathcal{F}$ from the input data. Then, the \ourencoder~(\ie, a series of the \ourblocks) augments the local context of multiple point cloud resolutions. Generally, the \ourblock~requires both semantic and geometric context as bilateral inputs. In particular, the first block inputs the preliminary $\mathcal{F}$ and the original 3D coordinates $\mathcal{P}$; while each of the others inputs its previous one's downsampled output and coordinates $\mathcal{P}$, as the semantic and geometric context respectively. Afterward, our \ourdecoder~(Sec.~\ref{sec:metho_fuse}) upsamples the \ourblocks' outputs, then adaptively fuses them as an output feature map. Finally, we predict semantic labels for all points via fully-connected layers.}
\label{fig:net}
\vspace{-3mm}
\end{figure*}
\vspace{1mm}
\noindent \textbf{Point Clouds Feature Representations:} 
Different from the individual point features in PointNet~\cite{qi2017pointnet}, the following methods focus on learning feature representations from local areas. Usually, the point neighbors are defined based on spatial metrics, \eg, 3D Euclidean distances in~\cite{qi2017pointnet++, liu2019relation, yan2020pointasnl, hu2020randla} or embedding similarities in~\cite{wang2019dynamic, qiu2019geometric, qiu2021dense}. By operating CNN-based modules over the neighborhoods, the local features of point clouds can be collected.

However, existing methods have limited capability to capture local details since they have not utilized the given information fully. Some works~\cite{qi2017pointnet, qi2017pointnet++, wang2019dynamic} only input the embedded features for each layer and lack the geometric restrictions in deep layers. Although current methods~\cite{liu2019relation,qiu2019geometric} employ local descriptors to strengthen the spatial relations, however, the additional computational cost is involved. The latest approaches~\cite{yan2020pointasnl, hu2020randla} combine the original 3D coordinates in all scales of the network, but the effect is subtle. Differently, we exploit the point features from two properties: the geometric and semantic contexts. By augmenting them in a bilateral fashion, we can synthesize an augmented local context to represent the point. 

\vspace{1mm}
\noindent \textbf{Semantic Segmentation Networks:} 2D semantic segmentation has been well studied in deep learning research. The basic FCN~\cite{long2015fully} applied a fully convolutional architecture to learn the features of each pixel. Further, UNet~\cite{ronneberger2015u} designed the symmetric downsampling and upsampling structure for image segmentation, while SegNet~\cite{badrinarayanan2017segnet} proposed the convolutional encoder-decoder structure. More recently, Chen~\etal~\cite{chen2020bidirectional} used a bi-directional gate to leverage multi-modality features, \ie, color and depth, for RGB-D images. 

In terms of 3D point clouds, most approaches are similar to the 2D image frameworks. For small-scale point clouds, the fully convolutional modules in~\cite{qi2017pointnet, wang2019dynamic, qiu2019geometric} are able to manage the complexity of the data. In contrast, for large-scale data, some networks~\cite{qi2017pointnet++, liu2019relation, hu2020randla, yan2020pointasnl} apply the convolutional encoder-decoders as SegNet~\cite{badrinarayanan2017segnet} does, to generate the point-wise representations. However, the performance may be less satisfactory: as lower resolutions are explored, it becomes more difficult to interpret the local context of the unstructured 3D points. Although methods~\cite{engelmann2020dilated, hu2020randla, qiu2021dense} attempt to tackle this problem by increasing the point's receptive field for a more detailed interpretation, it is expensive to find the optimal settings. Recent RandLA-Net~\cite{hu2020randla} achieves high efficiency using naive random sampling, while the network's accuracy and stability are sacrificed. Unlike the existing methods, we propose a bilateral augmentation structure to effectively process multi-resolution point clouds, and utilize an adaptive fusion method to represent the comprehensive point-wise features efficiently.
\section{Methodology}
\label{sec:metho}
A point cloud containing $N$ points can be described mainly from two aspects: 1) the inherent coordinates in 3D space $\mathcal{P}\in\mathbb{R}^{N\times3}$ which are explicitly obtained by 3D scanners indicating the geometric context of the points; and 2) the acquired features $\mathcal{F}\in\mathbb{R}^{N\times d}$ in $d$-dimensional feature space which can be implicitly encoded by CNN-based operations implying latent clues about semantic context. From this point of view, $\mathcal{P}$ and $\mathcal{F}$ are regarded as two properties of the point cloud features.

Although $\mathcal{P}$ is less informative for semantic analysis, it can enrich the basic perception of geometry for the network. On this front, we aim to fully utilize $\mathcal{P}$ and $\mathcal{F}$ in a reasonable way, which can support learning a comprehensive feature map for accurate semantic segmentation.

\subsection{\ourencoder}
\label{sec:metho_cross}
The \ourencoder~consists of a number of \ourblocks~to investigate the point cloud at different resolutions, as shown in Fig.~\ref{fig:net}. In the \ourblock, we intend to augment the local context of each point by involving the offsets that are mutually learned from the bilateral input information (\ie, $p_i\in\mathbb{R}^{3}$ and $f_i\in\mathbb{R}^{d}$), and then aggregate the augmented local context for the point feature representation. Particularly, we propose two novel units and a loss function to fulfill the intention.  

\vspace{1mm}
\noindent \textbf{Bilateral Augmentation:}
For a centroid $p_i$, we find its neighbors ${\forall}p_j\in Ni(p_i)$ using knn under the metric of 3D-Euclidean distance, while the corresponding neighbor features are denoted as $f_j$. To simultaneously capture both \emph{global} and \emph{local} information about the neighborhood, we combine the absolute position of the centroid and the relative positions of its neighbors as the local context ${\mathcal{G}_\psi}$. Accordingly, ${\mathcal{G}_\psi}(p_i) = [p_i; p_j-p_i]$ represents local geometric context in 3D space, while ${\mathcal{G}_\psi}(f_i) = [f_i; f_j-f_i]$ shows local semantic context in feature space.

However, ${\mathcal{G}_\psi}(p_i)$ and ${\mathcal{G}_\psi}(f_i)$ may be insufficient to represent the neighborhoods due to two reasons: 1) strict formation under a fixed constraint in 3D space could weaken the generalization capability of ${\mathcal{G}_\psi}$ in high-dimensional feature space, and 2) the ${\mathcal{G}_\psi}$ neighborhoods may have redundancy in the representations of close regions. To solve these issues and strengthen the generalization capability of the features, we can augment the local context by adding bilateral offsets, which shift the neighbors and densely affiliate them to the neighborhood's centroid.

To be specific, as the primary concern, we augment the local geometric context ${\mathcal{G}_\psi}(p_i)$ based on the rich semantic information of ${\mathcal{G}_\psi}(f_i)$. Particularly, we apply an MLP ($\mathcal{M}$) on ${\mathcal{G}_\psi}(f_i)$, to estimate the 3-DoF (Degrees of Freedom) bilateral offsets for the neighbors ${\forall}p_j\in Ni(p_i)$. Therefore, the shifted neighbors are formulated as:
\begin{equation}
\label{eq:shift}
    \tilde{p_j} = \mathcal{M}({\mathcal{G}_\psi}(f_i)) + p_j, \quad {\tilde{p_j}}\in\mathbb{R}^{3}.
\end{equation}
Afterwards, we gather the \emph{auxiliary} perception of the shifted neighbors to augment the local geometric context:
$
    {\tilde{\mathcal{G}_\psi}(p_i)} = [p_i; p_j-p_i; \tilde{p_j}];
$
where ${\tilde{\mathcal{G}_\psi}(p_i)}\in\mathbb{R}^{k\times 9}$ and $k$ is the number of neighbors.

Subsequently, the $d$-DoF bilateral offsets for the neighbor features $f_j$ can also be collected from ${\tilde{\mathcal{G}_\psi}(p_i)}$ since we expect the augmented local geometric context to further enhance the local semantic context. Similarly, the neighbor features are shifted as:
\begin{equation}
\label{eq:shift_feat}
    \tilde{f_j} = \mathcal{M}({\tilde{\mathcal{G}_\psi}(p_i)}) + f_j, \quad {\tilde{f_j}}\in\mathbb{R}^{d};
\end{equation}
and the augmented local semantic context is formed as: 
$
    {\tilde{\mathcal{G}_\psi}(f_i)} = [f_i; f_j-f_i; \tilde{f_j}],
$
where ${\tilde{\mathcal{G}_\psi}(f_i)}\in\mathbb{R}^{k\times 3d}.$ 

After further projecting the ${\tilde{\mathcal{G}_\psi}(p_i)}$ and ${\tilde{\mathcal{G}_\psi}(f_i)}$ by MLPs, we concatenate them as an augmented local context $\mathcal{G}_i$:
\begin{equation}
    \mathcal{G}_i = \mathrm{concat}\Big(\mathcal{M}\big({\tilde{\mathcal{G}_\psi}(p_i)}\big) , \mathcal{M}\big({\tilde{\mathcal{G}_\psi}(f_i)}\big)\Big)\in\mathbb{R}^{k\times d^{\prime}}.
\end{equation}

\noindent \textbf{Augmentation Loss:} 
We also introduce some penalties to regulate the learning process of the bilateral offsets in Eq.~\ref{eq:shift}. Since we should not only provide 3-DoF augmentation for the neighbors but also preserve the geometric integrity of a dense neighborhood, it is preferable to consider the neighbors as a whole instead of taking individual neighbors into account. Intuitively, we encourage the \emph{geometric center} of the shifted neighbors to approach the local centroid in 3D space by minimizing the $\ell_2$ distance:
\begin{equation}
\label{eq:pointloss}
    \mathcal{L}(p_i) = {\left\lVert \textstyle{\frac{1}{k}} \sum_{j=1}^{k} \tilde{p_j} - p_i \right\rVert}_2.
\end{equation}

\begin{algorithm}
\caption{Adaptive Fusion Module Pipeline}\label{alg:fusion}
\nonl\textbf{input:} $M$ multi-resolution feature maps $\{\mathcal{S}_1, \mathcal{S}_2, ..., \mathcal{S}_M\}$.\\
\nonl\textbf{output:} ${\mathcal{S}_{out}}$ for semantic segmentation.\\
\For{$\mathcal{S}_m\in \{\mathcal{S}_1, \mathcal{S}_2, ..., \mathcal{S}_M\}$}{
    \textbf{upsample:} $\tilde{\mathcal{S}_m}\leftarrow \mathcal{S}_m$\;
    \textbf{summarize:} $\phi_m\leftarrow \tilde{\mathcal{S}_m}$\;
}
\textbf{obtain:} ${\forall}\tilde{\mathcal{S}_m}\in \{\tilde{\mathcal{S}_1}, \tilde{\mathcal{S}_2}, ..., \tilde{\mathcal{S}_M}\}$, $\tilde{\mathcal{S}_m} \in\mathbb{R}^{N\times{c}}$;\\ 
\nonl and ${\forall}\phi_m\in \{\phi_1, \phi_2, ..., \phi_M\}$, $\phi_m \in\mathbb{R}^{N}$.\\
\textbf{regress:} $\{\Phi_1, \Phi_2, ..., \Phi_M\}\leftarrow \{\phi_1, \phi_2, ..., \phi_M\}$,\\
\nonl where $\Phi_m \in\mathbb{R}^{N}$.\\
\textbf{return:}\\
\nonl ${\mathcal{S}_{out}} = \sum_{m=1}^{M}{\Phi_m \times \tilde{\mathcal{S}_m}}$.\\
\end{algorithm}

\noindent \textbf{Mixed Local Aggregation:} Point-wise feature representation is crucial for the semantic segmentation task. Although non-parametric symmetric functions can efficiently summarize local information for the points, they cannot explicitly show the local distinctness, especially for close points sharing similar local context. To handle this problem, we propose a mixed local aggregation method to gather a precise neighborhood representation. 

Given the augmented local context $\mathcal{G}_i$, on the one hand, we directly collect the \emph{maximum} (prominent) feature from the $k$ neighbors for an overview of the neighborhood. On the other hand, we closely investigate the representations of the neighbors, refining and obtaining more details by learning the high-dimensional barycenter (\ie, weighted \emph{mean} point) over the neighborhood. In the end, we combine the two types of information, the local \emph{max} and \emph{mean} features, to precisely represent the point as:
\begin{equation}
\label{eq:mixed}
    s_i = \mathrm{concat}\Big(\max_{k}(\mathcal{G}_i), \underset{k,\Theta_i}{\mathrm{mean}}(\mathcal{G}_i)\Big),\quad {s_i}\in\mathbb{R}^{2d^\prime};
\end{equation}
where $\Theta_i$ is a set of learnable weights for $k$ neighbors. The implementation details are in Sec.~\ref{sec:impl_encoder}.

\subsection{Adaptive Fusion Module}
\label{sec:metho_fuse}
To efficiently analyze a real 3D scene consisting of a large number of points, we can gradually explore the point cloud in decreasing resolutions. Although it can be easily realized by applying the cascaded \ourblocks~for downsampled point cloud subsets, the corresponding output features become implicit and abstract. Therefore, it is essential to restore a feature map providing the original number of points and comprehensively interpret each point's encoded information. Specifically, we choose to fuse fine-grained representations from the multi-resolution feature maps adaptively.

Assume that $M$ lower resolutions of the point cloud are processed by the \ourencoder~(\ie, a cascaded set of the \ourblocks~as shown in Fig.~\ref{fig:net}), we extract a set of multi-resolution feature maps as $\{\mathcal{S}_1, \mathcal{S}_2, ..., \mathcal{S}_M\}$ including $\{N_1, N_2, ..., N_M\}$ points, respectively.\footnote{$N>N_1>N_2>...>N_M$, $N$ is the original size of a point cloud.} As claimed in Alg.~\ref{alg:fusion}, for each extracted feature map ${\forall}\mathcal{S}_m\in \{\mathcal{S}_1, \mathcal{S}_2, ..., \mathcal{S}_M\}$, we conduct progressive upsampling until a full-sized representation for all $N$ points is generated. Following a similar process, we reconstruct the full-sized feature maps $\{\tilde{\mathcal{S}_1}, \tilde{\mathcal{S}_2}, ..., \tilde{\mathcal{S}_M}\}$.

Although we manage to interpret the whole point cloud, in terms of each point, the upsampled feature representations that originate from multiple resolutions may result in different scales of information. To integrate the information and refine the useful context for semantic segmentation, we fuse the full-sized feature maps adaptively at point-level.

To be concrete, we additionally summarize the point-level information $\phi_m \in\mathbb{R}^{N}$ during the upsampling process of each full-sized feature map's generation, in order to capture basic point-level understanding from different scales. Next, by analyzing those point-level perceptions $\{\phi_1, \phi_2, ..., \phi_M\}$ on the whole, we regress the fusion parameters $\{\Phi_1,\Phi_2,..,\Phi_M\}$ corresponding to the full-sized feature maps $\{\tilde{\mathcal{S}_1}, \tilde{\mathcal{S}_2}, ..., \tilde{\mathcal{S}_M}\}$, respectively. In the end, a comprehensive feature map $\mathcal{S}_{out}$ for semantic segmentation is adaptively fused throughout multi-resolution features \wrt each point. Theoretically, it follows:
\begin{equation}
\label{eq:fusion}
    {\mathcal{S}_{out}} = \sum_{m=1}^{M}{\Phi_m \times \tilde{\mathcal{S}_m}}, \quad \Phi_m \in\mathbb{R}^{N}.
\end{equation}
More details about the \ourdecoder~implementation are presented in Sec.~\ref{sec:impl_decoder}. 


\section{Implementation Details}
\label{sec:impl}
Based on the key structures in Sec.~\ref{sec:metho}, we form an effective network for the semantic segmentation of real point clouds scenes. As illustrated in Fig.~\ref{fig:net}, our network has three modules: the \ourextractor, the \ourencoder, and the \ourdecoder. We introduce the details of each module in the following sections.

\subsection{\ourextractor}
\label{sec:impl_extractor}
Besides spatial 3D coordinates, some datasets may include other clues, \eg, RGB colors, light intensity, \etc. To create an overall impression of the whole scene, initially, we apply the \ourextractor~to acquire preliminary semantic knowledge from all of the provided information. Given the advantages of an MLP that it can represent the features flexibly in a high-dimensional embedding space, empirically, we apply a single-layer MLP (\ie, a 1-by-1 convolutional layer followed by  batch normalization~\cite{ioffe2015batch} and an activation function like $\mathrm{ReLU}$) to obtain high-level compact features. Fig.~\ref{fig:net} shows the acquired features $\mathcal{F}$ from the \ourextractor~which are forwarded to the \ourencoder, along with the 3D coordinates $\mathcal{P}$.

\subsection{\ourencoder~Implementation}
\label{sec:impl_encoder}
As mentioned before, the \ourencoder~explores the different resolutions of point cloud data. For the sake of stability, we use CUDA-based Farthest Point Sampling (FPS) to sample the data based on its 3D distribution. Particularly, the \ourencoder~deploys cascaded \ourblocks~to gradually process the lower resolutions of the point cloud: \eg, $N{\rightarrow}{\frac{N}{4}}{\rightarrow}{\frac{N}{16}}{\rightarrow}{\frac{N}{64}}{\rightarrow}{\frac{N}{256}}$. Meanwhile, the dimensions of the outputs are increasing as: $32{\rightarrow}128{\rightarrow}256{\rightarrow}512{\rightarrow}1024$. In this regard, the behavior of the \ourencoder~processing the 3D point clouds is similar to the classical CNNs for 2D images, which extend the channel number while shrinking the image size for a concise description.

Inside each \ourblock, an efficient k-nearest neighbor using the nanoflann~\cite{blanco2014nanoflann} library speeds up neighbor searching in the bilateral augmentation unit. Empirically, we set k=12 for all experiments in this work. For the mixed local aggregation unit, the local \emph{max} feature is collected by operating a max-pooling function along the neighbors. Following a similar operation in~\cite{hu2020randla}, we simultaneously refine and re-weight the neighbors through a single-layer MLP and a $\mathrm{softmax}$ function, then aggregate the barycenter of local embeddings as the local \emph{mean} feature. Finally, the local \emph{max} and \emph{mean} features are concatenated as the output of the mixed local aggregation unit.

\subsection{\ourdecoder~Implementation}
\label{sec:impl_decoder}
As explained in Sec.~\ref{sec:metho_fuse}, our \ourdecoder~aims to upsample the multi-resolution outputs of the \ourencoder, and then adaptively fuse them as a comprehensive feature map for the whole point cloud scene. To be more specific with the upsampling process, at first, a single-layer MLP integrates the channel-wise information of the output feature maps. Then, we point-wisely interpolate a higher-resolution feature map using nearest neighbor interpolation~\cite{keys1981cubic}, since it is more efficient for large-scale data than Feature Propagation~\cite{qi2017pointnet++} that requires huge computational cost for neighbors and weights. Moreover, we symmetrically attach the features from the same resolution in order to increase diversity and distinctness for nearby points. Finally, a higher-resolution feature map is synthesized via another single-layer MLP.

The upsampling process is continuously performed to get full-sized feature maps $\{\tilde{\mathcal{S}_1}, \tilde{\mathcal{S}_2}, ..., \tilde{\mathcal{S}_M}\}$ from the multi-resolution outputs of the \ourencoder. During this process, we also use a fully-connected layer to summarize the point-level information $\phi_m$ once a full-sized feature map $\tilde{\mathcal{S}_m}$ is reconstructed. To analyze the summarized information, we concatenate $\{\phi_1, \phi_2, ..., \phi_M\}$, and point-wisely normalize them using $\mathrm{softmax}$. As a result, the fusion parameters $\{\Phi_1, \Phi_2, ..., \Phi_M\}$ are adaptively regressed \wrt each point. After calculating a weighted sum of the upsampled feature maps (Eq.~\ref{eq:fusion}), we eventually combine a feature map containing all points for whole scene semantic segmentation. Besides, a structure chart of this module is provided in the supplementary material.

\subsection{Loss Function}
Using the fused output of the \ourdecoder, the FC layers predict the confidence scores for all candidate semantic classes. Generally, cross-entropy loss $\mathcal{L}_{CE}$ is computed for back-propagation. Further, we include point-level augmentation losses $\mathcal{L}(p_i)$ that are formulated following Eq.~\ref{eq:pointloss}. In terms of a \ourblock~processing $N_m$ points, the total augmentation loss regarding $N_m$ points would be $\mathcal{L}_{m}=\sum_{i=1}^{N_m} \mathcal{L}(p_i)$. Hence, for our network containing $M$ \ourblocks, the overall loss is:
\begin{equation}
\label{equ:all_loss}
    \mathcal{L}_{all} = \mathcal{L}_{CE} + \sum_{m=1}^{M} \omega_m \cdot \mathcal{L}_{m},
\end{equation}
where $\omega_m$ is a hyper-parameter of weight for each~\ourblock.
\section{Experiments}
\label{sec:exp}
\subsection{Experimental Settings}
\noindent \textbf{Datasets:}
In this work, we are targeting the semantic segmentation of real point cloud scenes. To validate our approach, we conduct experiments on three 3D benchmarks, which present different scenes in the real world. 
\begin{itemize}
 \item \textbf{S3DIS:} 
Stanford Large-Scale 3D Indoor Spaces (S3DIS)~\cite{armeni2017joint} dataset is collected from \emph{indoor} working environments. In general, there are six sub-areas in the dataset, each containing $\sim$50 different rooms. The number of points in most rooms varies from 0.5 million to 2.5 million, depending on the room's size. All points are provided with both 3D coordinates and color information and labeled as one of 13 semantic categories. We adopt a 6-fold strategy~\cite{qi2017pointnet} for evaluation.
\begin{table}
\begin{center}
\captionsetup{font=small, skip=3pt}
\caption{Semantic segmentation (6-fold cross-validation) results (\%) on the \emph{S3DIS} dataset~\cite{armeni2017joint}. (\textbf{mAcc}: average class accuracy, \textbf{OA}: overall accuracy, \textbf{mIoU}: mean Intersection-over-Union. \enquote{-} indicates unknown result.)}
\resizebox{0.9\columnwidth}{!}{
\begin{tabular}{|c|c|c c |c|c|}
\hline
year     &{Method}        & \textbf{mAcc}  & \;\;\textbf{OA}\;\;  & \textbf{mIoU} \\ \hline
\multirow{2}{*}{2017}&PointNet \cite{qi2017pointnet}   & 66.2 & 78.6 & 47.6         \\ 
&PointNet++ \cite{qi2017pointnet++}  & 67.1 & 81.0 & 54.5         \\\hline
\multirow{4}{*}{2018}&A-SCN \cite{xie2018attentional}    & - & 81.6 & 52.7          \\
&PointCNN \cite{li2018pointcnn}   & 75.6 & 88.1 & 65.4          \\
&SPG \cite{landrieu2018large}    & 73.0 & 85.5 & 62.1          \\\hline
\multirow{5}{*}{2019}
&DGCNN \cite{wang2019dynamic}    & - & 84.1 & 56.1         \\ 
&KP-Conv \cite{thomas2019kpconv}      & 79.1 & - & 70.6  \\
&ShellNet \cite{zhang2019shellnet}     & - & 87.1 & 66.8  \\
&PointWeb \cite{zhao2019pointweb}    & 76.2 & 87.3 & 66.7  \\
&SSP+SPG \cite{landrieu2019point}     & 78.3 & 87.9 & 68.4  \\\hline
\multirow{6}{*}{2020}
&Seg-GCN \cite{lei2020seggcn}     & 77.1 & 87.8 & 68.5  \\
&PointASNL \cite{yan2020pointasnl}   & 79.0 &88.8 & 68.7  \\
&RandLA-Net \cite{hu2020randla}      & 82.0 & 88.0 & 70.0        \\
&MPNet \cite{he2020learning}      & - & 86.8 & 61.3        \\
&InsSem-SP \cite{liu2020self}     & 74.3 & 88.5 & 64.1        \\
 \cline{2-5} 
&\textbf{Ours}  & \textbf{83.1} & \textbf{88.9} & \textbf{72.2}  \\ \hline
\end{tabular}
\label{tab:s3dis}
}
\end{center}
\vspace{-5mm}
\end{table}

\begin{table*}
\begin{center}
\captionsetup{skip=3pt}
\caption{Semantic segmentation (semantic-8) results (\%) on the \emph{Semantic3D} dataset~\cite{hackel2017semantic3d}.}
\rowcolors{2}{gray!20}{}
\resizebox{0.75\textwidth}{!}{
\begin{tabular}{c c c c c c c c c c c}
\Xhline{3\arrayrulewidth}
\multirow{2}{*}{Method}&\multirow{2}{*}{\textbf{OA}} &\multirow{2}{*}{\textbf{mIoU}}  &man-made&natural&high&low&\multirow{2}{*}{buildings}&hard&scanning&\multirow{2}{*}{cars}  \\
& &  &terrain&terrain&vegetation&vegetation&&scape&artefacts&   \\\hline
TMLC-MS~\cite{hackel2016fast} &85.0&49.4&91.1&69.5&32.8&21.6&87.6&25.9&11.3&55.3\\
EdgeConv-PN~\cite{contreras2019edge} &89.4&61.0&91.2&69.8&51.4&58.5&90.6&33.0&24.9&68.6\\
PointNet++~\cite{qi2017pointnet++} &85.7&63.1&81.9&78.1&64.3&51.7&75.9&36.4&43.7&72.6\\
SnapNet~\cite{boulch2018snapnet} &91.0&67.4&89.6&79.5&74.8&56.1&90.9&36.5&34.3&77.2\\
PointConv~\cite{wu2019pointconv} &91.8&69.2&92.2&79.2&73.1&62.7&92.0&28.7&43.1&82.3\\
PointGCR~\cite{ma2020global} &92.1&69.5&93.8&80.0&64.4&66.4&93.2&39.2&34.3&85.3\\
PointConv-CE~\cite{9136884} &92.3&71.0&92.4&79.6&72.7&62.0&93.7&40.6&44.6&82.5\\
RandLA-Net~\cite{hu2020randla} &94.2&71.8&96.0&88.6&65.3&62.0&\textbf{95.9}&49.8&27.8&89.3\\
SPG~\cite{landrieu2018large} &92.9&\textbf{76.2}&91.5&75.6&\textbf{78.3}&\textbf{71.7}&94.4&\textbf{56.8}&\textbf{52.9}&88.4\\\hline
\textbf{Ours}  &\textbf{94.9}&75.4&\textbf{97.9}&\textbf{95.0}&70.6&63.1&94.2&41.6&50.2&\textbf{90.3}   \\ \Xhline{3\arrayrulewidth}
\end{tabular}
\label{tab:semantic3d}
}
\end{center}
\vspace{-3mm}
\end{table*}

\begin{table*}
\begin{center}
\captionsetup{skip=3pt}
\caption{Semantic segmentation (single-scan) results (\%) on the \emph{SemanticKITTI} dataset~\cite{behley2019semantickitti}.}
\rowcolors{4}{gray!20}{} 
\resizebox{0.98\textwidth}{!}{
\begin{tabular}{c c c c c c c c c c c c c c c c c c c c c}
\Xhline{3\arrayrulewidth}
 &&&\multirow{4}{*}{\rotatebox[origin=r]{90}{sidewalk}}&\multirow{4}{*}{\rotatebox[origin=r]{90}{parking}}&\multirow{4}{*}{\rotatebox{90}{\small{other-ground}}}&\multirow{4}{*}{\rotatebox[origin=r]{90}{building}}&&&&\multirow{4}{*}{\rotatebox{90}{\small{motorcycle}}}&\multirow{4}{*}{\rotatebox{90}{\small{other-vehicle}}}&\multirow{4}{*}{\rotatebox[origin=r]{90}{\small{\quad vegetation}}}& &&&\multirow{4}{*}{\rotatebox[origin=r]{90}{bicyclist}}&\multirow{4}{*}{\rotatebox{90}{\small{motorcyclist}}}&&&\multirow{4}{*}{\rotatebox{90}{traffic-sign}}\\
\multirow{4}{*}{\large Method} &\multirow{4}{*}{\large\textbf{mIoU}}&\multirow{4}{*}{\rotatebox[origin=c]{90}{\;road}}&&&&&\multirow{4}{*}{\rotatebox[origin=r]{90}{\;car}}&\multirow{4}{*}{\rotatebox[origin=r]{90}{\quad truck}}&\multirow{4}{*}{\rotatebox[origin=c]{90}{\quad\; bicycle}}&&&&\multirow{4}{*}{\rotatebox[origin=r]{90}{\quad trunk}}&\multirow{4}{*}{\rotatebox[origin=c]{90}{\quad terrain}}&\multirow{4}{*}{\rotatebox[origin=c]{90}{\quad person}}&&&\multirow{4}{*}{\rotatebox[origin=r]{90}{\quad fence}}&\multirow{4}{*}{\rotatebox[origin=c]{90}{\;pole}}&\\
&&&&&&&&&&&&&&&&&&&&\\
&&&&&&&&&&&&&&&&&&&& \\\hline
PointNet~\cite{qi2017pointnet} &14.6&61.6&35.7&15.8&1.4&41.4&46.3&0.1&1.3&0.3&0.8&31.0&4.6&17.6&0.2&0.2&0.0&12.9&2.4&3.7\\
PointNet++~\cite{qi2017pointnet++} &20.1&72.0&41.8&18.7&5.6&62.3&53.7&0.9&1.9&0.2&0.2&46.5&13.8&30.0&0.9&1.0&0.0&16.9&6.0&8.9\\
SquSegV2~\cite{wu2019squeezesegv2} &39.7&88.6&67.6&45.8&17.7&73.7&81.8&13.4&18.5&17.9&14.0&71.8&35.8&60.2&20.1&25.1&3.9&41.1&20.2&36.3\\
TangentConv~\cite{tatarchenko2018tangent} &40.9&83.9&63.9&33.4&15.4&83.4&90.8&15.2&2.7&16.5&12.1&79.5&49.3&58.1&23.0&28.4&8.1&49.0&35.8&28.5\\
PointASNL~\cite{yan2020pointasnl} &46.8&87.4&74.3&24.3&1.8&83.1&87.9&39.0&0.0&25.1&29.2&84.1&52.2&\textbf{70.6}&34.2&\textbf{57.6}&0.0&43.9&57.8&36.9\\
RandLA-Net~\cite{hu2020randla} &53.9&90.7&73.7&60.3&20.4&86.9&94.2&40.1&26.0&25.8&38.9&81.4&61.3&66.8&49.2&48.2&7.2&56.3&49.2&47.7\\
PolarNet~\cite{zhang2020polarnet} &54.3&90.8&74.4&61.7&21.7&90.0&93.8&22.9&40.3&30.1&28.5&84.0&65.5&67.8&43.2&40.2&5.6&67.8&51.8&57.5\\
MinkNet42~\cite{choy20194d} &54.3 &91.1 &69.7 &63.8 &29.3 &\textbf{92.7} &94.3 &26.1 &23.1 &26.2 &36.7 &83.7 &68.4 &64.7 &43.1 &36.4 &7.9 &57.1 &57.3 &60.1\\
FusionNet~\cite{zhang2020deep} &\textbf{61.3} &\textbf{91.8} &\textbf{77.1} &\textbf{68.8} &\textbf{30.8} &92.5 &95.3 &41.8 &\textbf{47.5} &\textbf{37.7} &34.5 &\textbf{84.5} &\textbf{69.8} &68.5 &\textbf{59.5} &56.8 &11.9 &\textbf{69.4} &\textbf{60.4} &\textbf{66.5}\\
\hline
\textbf{Ours}  &59.9&90.9&74.4&62.2&23.6&89.8&\textbf{95.4}&\textbf{48.7}&31.8&35.5&\textbf{46.7}&82.7&63.4&67.9&49.5&55.7&\textbf{53.0}&60.8&53.7&52.0   \\ \Xhline{3\arrayrulewidth}
\end{tabular}
\label{tab:kitti}
}
\end{center}
\vspace{-3mm}
\end{table*}
 
 \item \textbf{Semantic3D:}  
The points in Semantic3D~\cite{hackel2017semantic3d} are scanned in \emph{natural} scenes depicting various rural and urban views. Overall, this dataset contains more than four billion points manually marked in eight semantic classes. In particular, the dataset has two test sets for online evaluation: the full test set (\ie, semantic-8) has 15 scenes with over 2 billion points, while its subset (\ie, reduced-8) has four selected scenes with $\sim$0.1 billion sampled points. In this work, we use both 3D positions and colors of points for training and then infer the dense scenes of entire \emph{semantic-8} test set. 

 \item \textbf{SemanticKITTI:} 
SemanticKITTI~\cite{behley2019semantickitti} was introduced based on the well-known KITTI Vision~\cite{Geiger2012CVPR} benchmark illustrating complex \emph{outdoor} traffic scenarios. There are 22 stereo sequences, which are densely recorded as scans ($\sim$0.1 million points in each scan) and precisely annotated in 19 semantic classes. Particularly, 11 of the 22 sequences are provided with labels, while the results of the other ten sequences (over 20k scans) are for online evaluation. As in~\cite{behley2019semantickitti}, we take sequence 08 as the validation set, while the remaining ten labeled sequences ($\sim$19k scans) are for training.
\end{itemize}

\vspace{1mm}
\noindent \textbf{Training Settings:}
We train for 100 epochs on a single GeForce RTX 2080Ti GPU with a batch size between 4 to 6, depending on the amount of input points (about $40\times 2^{10}$ to $64\times 2^{10}$) for different datasets. In addition, the Adam~\cite{kingma2014adam} optimizer is employed to minimize the overall loss in Eq.~\ref{equ:all_loss}; the learning rate starts from 0.01 and decays with a rate of 0.5 after every 10 epochs. We implement the project\footnote{The codes and test results are available at \url{https://github.com/ShiQiu0419/BAAF-Net}.} in Python and Tensorflow~\cite{abadi2016tensorflow} platforms using Linux.

\vspace{1mm}
\noindent \textbf{Evaluation Metrics:}
To evaluate our semantic segmentation performance, we largely use the mean Intersection-over-Union (mIoU), the average value of IoUs for all semantic classes upon the whole dataset. Further, we also provide the overall accuracy (OA) regarding all points and the average class accuracy (mAcc) of all semantic classes. As for S3DIS~\cite{armeni2017joint}, we compute the mIoU based on all predicted sub-areas following the 6-fold strategy. Similarly, for both Semantic3D~\cite{hackel2017semantic3d} and SemanticKITTI~\cite{behley2019semantickitti}, we provide the online submission testing results of general mIoU and OA, as well as the IoU for each semantic category.

\begin{figure*}
\begin{center}
\includegraphics[width=0.98\textwidth]{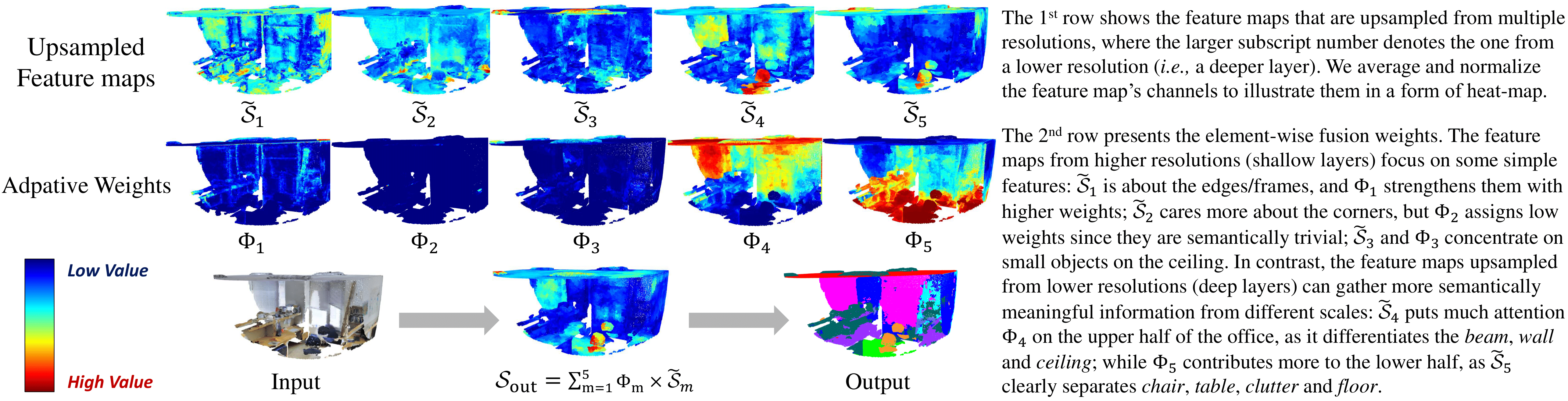}
\end{center}
\vspace{-5mm}
\captionsetup{font=small}
\caption{Behavior analysis of the Adaptive Fusion Module (Sec.~\ref{sec:metho_fuse}) based on an office scene in \emph{S3DIS} dataset. By fusing the upsampled feature maps in a simple but adaptive way, we aggregate the advantages from different scales, and generate $\mathcal{S}_{out}$ for semantic segmentation.}
\label{fig:visual}
\vspace{-3mm}
\end{figure*}

\begin{table}
\begin{center}
\captionsetup{font=small, skip=6pt}
\caption{Ablation studies about the \ourblock~testing on Area 5, \emph{S3DIS} dataset. ($\tilde{p_i}\rightarrow \tilde{f_i}$: learn 3-DoF offsets $\tilde{p_i}$ first and $d$-DoF offsets $\tilde{f_i}$ afterwards as per Sec.~\ref{sec:metho_cross}; $\tilde{f_i}\rightarrow \tilde{p_i}$: learn $\tilde{f_i}$, and then $\tilde{p_i}$; $\mathcal{L}(\cdot)$: calculate augmentation loss as per Eq.~\ref{eq:pointloss}; \emph{mixed}: mixed local aggregation following Eq.~\ref{eq:mixed}; \emph{max}: local max feature $\max_{k}(\mathcal{G}_i)$ only; \emph{mean}: local mean feature $\underset{k,\Theta_i}{\mathrm{mean}}(\mathcal{G}_i)$ only.)}
\vspace{-2mm}
\resizebox{0.9\columnwidth}{!}{
\begin{tabular}{c|ccc|c}
\Xhline{3\arrayrulewidth}
\multirow{2}{*}{Model}&bilateral&augmentation&local&\multirow{2}{*}{mIoU} \\
&offsets&loss&aggregation&\\\hline\xrowht{7pt}
$\mathrm{B}_0$ &none&none&max& 61.8         \\\xrowht{7pt}
$\mathrm{B}_1$ &$\tilde{f_i}\rightarrow \tilde{p_i}$&$\mathcal{L}(f_i)$&mixed& 64.2       \\\xrowht{7pt}
$\mathrm{B}_2$ &$\tilde{p_i}\rightarrow \tilde{f_i}$&$\mathcal{L}(p_i)+\mathcal{L}(f_i)$&mixed& 64.3          \\\xrowht{7pt}
$\mathrm{B}_3$ &$\tilde{p_i}\rightarrow \tilde{f_i}$&none&mixed& 64.2          \\\xrowht{7pt}
$\mathrm{B}_4$ &$\tilde{p_i}\rightarrow \tilde{f_i}$&$\mathcal{L}(p_i)$&max&  64.6         \\\xrowht{7pt}
$\mathrm{B}_5$ &$\tilde{p_i}\rightarrow \tilde{f_i}$&$\mathcal{L}(p_i)$&mean&  64.8        \\ \hline \xrowht{12pt}
$\bm{\mathrm{B}_6}$   &$\tilde{p_i}\rightarrow \tilde{f_i}$&$\mathcal{L}(p_i)$&mixed   & \textbf{65.4}  \\\Xhline{3\arrayrulewidth}      
\end{tabular}
\label{tab:cross-modal}
}
\end{center}
\vspace{-6mm}
\end{table}

\subsection{Semantic Segmentation Results}
\noindent \textbf{S3DIS:}
Tab.~\ref{tab:s3dis} quantitatively presents the performance of our network on the S3DIS dataset compared with other state-of-the-art methods. Notably, although recent methods achieve good results regarding overall accuracy, this metric is unable to indicate the semantic segmentation ability due to class imbalance among different categories. In general, we significantly outperform the competitors regarding the metrics of average class accuracy (83.1\%) and mIoU (72.2\%). Moreover, we visualize the Adaptive Fusion Module's upsampled features maps and adaptive weights in Fig.~\ref{fig:visual} (better in a zoom-in and colored view) based on S3DIS, in order to intuitively analyze the module's behavior while fusing the multi-resolution feature maps.  

\vspace{1mm}
\noindent \textbf{Semantic3D:} 
We also perform well on the natural views of the Semantic3D dataset. As Tab.~\ref{tab:semantic3d} indicates, we surpass other methods in three out of the eight classes; and our method is accurate on three categories, \ie, \emph{human-made and natural terrains, cars}, whose IoUs are all higher than 90\%. Considering the results of both overall accuracy (94.9\%) and mIoU (75.4\%) upon two billion testing points, our method accurately classifies the semantic labels of points in real scenes, especially for large-scale data.

\begin{table}
\begin{center}
\captionsetup{font=small, skip=5pt}
\caption{Ablation studies about the Adaptive Fusion Module testing on Area 5, \emph{S3DIS} dataset. ($\{\tilde{\mathcal{S}_m}\}$: a set of upsampled feature maps, $\tilde{\mathcal{S}_1}$,..,$\tilde{\mathcal{S}_M}$, as mentioned in Alg.~\ref{alg:fusion}; concat, $\sum$ and $\prod$: the concatenation, element-wise sum and element-wise multiplication for the set $\{\tilde{\mathcal{S}_m}\}$; $\{\Psi_m\}$: scalars for the set $\{\tilde{\mathcal{S}_m}\}$; $\{\Phi_m\}$: point-level fusion parameters as explained in Sec.~\ref{sec:metho_fuse} and~\ref{sec:impl_decoder}.)}
\resizebox{0.9\columnwidth}{!}{
\begin{tabular}{c|ccc|c}
\Xhline{3\arrayrulewidth}
\multirow{2}{*}{Model}&upsampled&fusion&\multirow{2}{*}{$\mathcal{S}_{out}$}&\multirow{2}{*}{mIoU} \\
&feature map&parameters&&\\\hline\xrowht{8pt}
$\mathrm{A}_0$ &$\tilde{\mathcal{S}_M}$&none&$\tilde{\mathcal{S}_M}$&64.1        \\\xrowht{8pt}
$\mathrm{A}_1$ &$\{\tilde{\mathcal{S}_m}\}$&none&$\sum {\tilde{\mathcal{S}_m}}$& 64.7          \\\xrowht{8pt}
$\mathrm{A}_2$ &$\{\tilde{\mathcal{S}_m}\}$&none&$\prod {\tilde{\mathcal{S}_m}}$& 64.2         \\\xrowht{8pt}
$\mathrm{A}_3$ &$\{\tilde{\mathcal{S}_m}\}$&none&$\mathrm{concat}(\{\tilde{\mathcal{S}_m}\})$& 65.1        \\\xrowht{8pt}
$\mathrm{A}_4$ &$\{\tilde{\mathcal{S}_m}\}$&$\{\Psi_m\}$&$\sum {\Psi_m \times \tilde{\mathcal{S}_m}}$& 65.1         \\ \hline \xrowht{11pt}
$\bm{\mathrm{A}_5}$   &$\{\tilde{\mathcal{S}_m}\}$&$\{\Phi_m\}$&$\sum {\Phi_m \times \tilde{\mathcal{S}_m}}$   & \textbf{65.4}  \\\Xhline{3\arrayrulewidth}      
\end{tabular}
\label{tab:fusion}
}
\end{center}
\vspace{-7mm}
\end{table}
\vspace{1mm}
\noindent \textbf{SemanticKITTI:} 
Although SemanticKITTI is challenging due to the complex scenarios in traffic environments, our network can effectively identify the semantic labels of points. As shown in Tab.~\ref{tab:kitti}, we exceed other advanced approaches in 4 of all 19 classes. Particularly, we perform well regarding the \emph{small} objects in dense scans such as \emph{car, truck, other-vehicle, motorcyclist}, \etc The outstanding results can be credited to our point-level adaptive fusion method, which thoroughly integrates the different scales. Overall, our network boosts a lot (5.6\% mIoU) compared to the latest point and grid-based methods~\cite{yan2020pointasnl, hu2020randla, zhang2020polarnet}, while is slightly behind the state-of-the-art work~\cite{zhang2020deep} using sparse tensor-based framework~\cite{choy20194d}. As our main ideas of bilateral augmentation and adaptive fusion are fairly adaptable, more experiments with different frameworks will be studied in the future. 

\begin{table}
\begin{center}
\captionsetup{font=small, skip=3pt}
\caption{Complexity analysis of different semantic segmentation networks on \emph{SemanticKITTI}. (\enquote{-} indicates the unknown result.)}
\resizebox{0.95\columnwidth}{!}{
\begin{tabular}{c|ccc|c}
\Xhline{3\arrayrulewidth}
\multirow{2}{*}{Method}&Parameters&Max Capacity&Inference Speed&\multirow{2}{*}{mIoU} \\
&(\emph{millions})&(\emph{million points})&(\emph{scans/second})&\\\hline\xrowht{7pt}
PointNet~\cite{qi2017pointnet}        & 0.8& 0.49& 21.2            & 14.6         \\\xrowht{7pt}
PointNet++~\cite{qi2017pointnet++}       & 0.97  &  0.98& 0.4             & 20.1      \\\xrowht{7pt}
SPG~\cite{landrieu2018large}     & \textbf{0.25}&  -&  0.1           & 17.4      \\\xrowht{7pt}
RandLA-Net~\cite{hu2020randla}       & 1.24&\textbf{1.03} & \textbf{22}                & 53.9      \\\hline\xrowht{10pt}
\textbf{Ours}       & 1.23& 0.9 & 4.8   & \textbf{59.9}  \\\Xhline{3\arrayrulewidth}      
\end{tabular}
\label{tab:complexity}
}
\end{center}
\vspace{-7mm}
\end{table}
\subsection{Ablation Studies}
\label{sec:exp_abl}
\noindent \textbf{\ourblock:}
In Tab.~\ref{tab:cross-modal}, we study the \ourblock's structure by investigating the components individually. $\mathrm{B}_0$ is the baseline model which only max-pools the concatenation of the basic local geometric $\mathcal{G}_{\psi}(p_i)$ and semantic context $\mathcal{G}_{\psi}(f_i)$; while rest models use different components based on the same structure of bilateral augmentation. From model $\mathrm{B}_1$\&$\mathrm{B}_2$, we observe that the semantic augmentation loss $\mathcal{L}(f_i)$ has no effect since augmenting the semantic features in embedding space is implicit. In contrast, the bilateral offsets $\tilde{p_i}$ with the geometric augmentation loss $\mathcal{L}(p_i)$ improves a bit (model $\mathrm{B}_4$\&$\mathrm{B}_5$). Taking the advantages from both local \emph{max} and \emph{mean} features, we conclude that the best form of the \ourblock~is using \emph{mixed} local aggregation ($\mathrm{B}_6$).   

\vspace{1mm}
\noindent \textbf{Adaptive Fusion Module:}
In Tab.~\ref{tab:fusion}, by comparing models $\mathrm{A}_1$, $\mathrm{A}_2$\&$\mathrm{A}_3$ with the baseline $\mathrm{A}_0$ that only upsamples the final output of the \ourencoder, we notice that utilizing the upsampled features maps that originate from multiple resolutions can benefit the performance. However, the fusion method decides whether the effects are significant or not: regular summation ($\mathrm{A}_1$) or multiplication ($\mathrm{A}_2$) is not desirable, while concatenation ($\mathrm{A}_3$) contributes more to the final prediction. For a general fusion ($\mathrm{A}_4$) \wrt each feature map, we regress a set of scalars $\{\Psi_m\}$ based on the squeezed information~\cite{hu2018squeeze} of the feature maps. Instead, a more flexible fusion operating adaptively at point-level ($\mathrm{A}_5$) achieves better results since semantic segmentation relies more on point-wise feature representations.  

\vspace{1mm}
\noindent \textbf{Network Complexity:}
Network complexity is essential to the practical application of point clouds. In Tab.~\ref{tab:complexity}, we use similar metrics as~\cite{hu2020randla} to study the inference using the trained models. The complexity and capacity (\ie, the number of parameters, and the maximum number of points for prediction) of our model are comparable to~\cite{qi2017pointnet++, hu2020randla}. Although \cite{hu2020randla}~is efficient for one-time inference, they require multiple evaluations to minimize the impact of random sampling, while we obtain more effective and stable semantic segmentation results in different real scenes such as the examples shown in Fig.~\ref{fig:compare}. More visualizations and experimental results are presented in the supplementary material.
\section{Conclusions}
\label{sec:concl}
This paper focuses on fundamental analysis and semantic segmentation for real point clouds scenes. Specifically, we propose a network leveraging the ideas of augmenting the local context bilaterally and fusing multi-resolution features for each point adaptively. Particularly, we achieve outstanding performance on three benchmarks, including S3DIS, Semantic3D, and SemanticKITTI. Further, we analyze the modules' properties by conducting related ablation studies, and intuitively visualize the network's effects. In the future, we expect to optimize the efficiency for real-time applications, exploit the key ideas in different frameworks, and promote the primary structures for more 3D tasks such as object detection, instance segmentation, \etc

{\small
\bibliographystyle{ieee_fullname}
\bibliography{egbib}
}
\clearpage
\appendix
\noindent\textbf{\Large{Supplementary Material}}
\section{Overview}
This supplementary material provides more network details, experimental results, and visualizations of our semantic segmentation results.
\section{Network Details}
In Fig.~\ref{fig:net} of the main paper, we present the general architecture of our semantic segmentation network as well as the structure of the \ourblock. In this section, we provide more details about the different components of our network.

\subsection{Key Modules}
\noindent \textbf{\ourextractor:}  As stated, we apply a single-layer MLP containing eight 1$\times$1 kernels to extract the semantic context $\mathcal{F}$ from the input information $\mathcal{I}\in\mathbb{R}^{N\times C_{in}}$, where $N$ is the number of input points. Hence, $\mathcal{F}$ is acquired as:
$$
\begin{aligned}
    \mathcal{F} &= \mathrm{ReLU}\Big(\mathrm{BN}\big(\mathrm{Conv}_{1\times1}^{8}(\mathcal{I})\big)\Big), \quad \mathcal{F}\in\mathbb{R}^{N\times 8};
\end{aligned}
$$
where $\mathrm{Conv}$ denotes a convolution layer whose subscript is the kernel size, and the superscript is the number of kernels. $\mathrm{BN}$ represents a batch normalization layer, while $\mathrm{ReLU}$ is a ReLU activation layer. Later on, $\mathcal{F}$ is forwarded to the \ourencoder, together with the 3D coordinates $\mathcal{P}\in\mathbb{R}^{N\times 3}$.

\vspace{3mm}
\noindent \textbf{\ourencoder:}  In practice, we apply five \ourblocks~with Farthest Point Sampling (FPS) to realize the \ourencoder~($\mathcal{B}$). Using the same annotations of the main paper’s Sec.~\ref{sec:impl_encoder}, the extracted multi-resolution feature maps are:
$$
    \{{\mathcal{S}_1}, {\mathcal{S}_2}, {\mathcal{S}_3}, {\mathcal{S}_4}, {\mathcal{S}_5}\} = \mathcal{B}(\mathcal{P}, \mathcal{F});
$$
where:
$$
\mathcal{S}_1\in\mathbb{R}^{{\frac{N}{4}}\times 32}, \quad \mathcal{S}_2\in\mathbb{R}^{{\frac{N}{16}}\times 128}, \quad \mathcal{S}_3\in\mathbb{R}^{{\frac{N}{64}}\times 256},
$$
$$
 \mathcal{S}_4\in\mathbb{R}^{{\frac{N}{256}}\times 512}, \quad \mathcal{S}_5\in\mathbb{R}^{{\frac{N}{512}}\times 1024}.
$$

Particularly, the downsampling ratios and feature dimensions are simply adopted from~\cite{hu2020randla}, since we mainly focus on the structure design rather than fine-tuning the hyper-parameters in this work.
\begin{figure}
\begin{center}
\includegraphics[width=0.98\columnwidth]{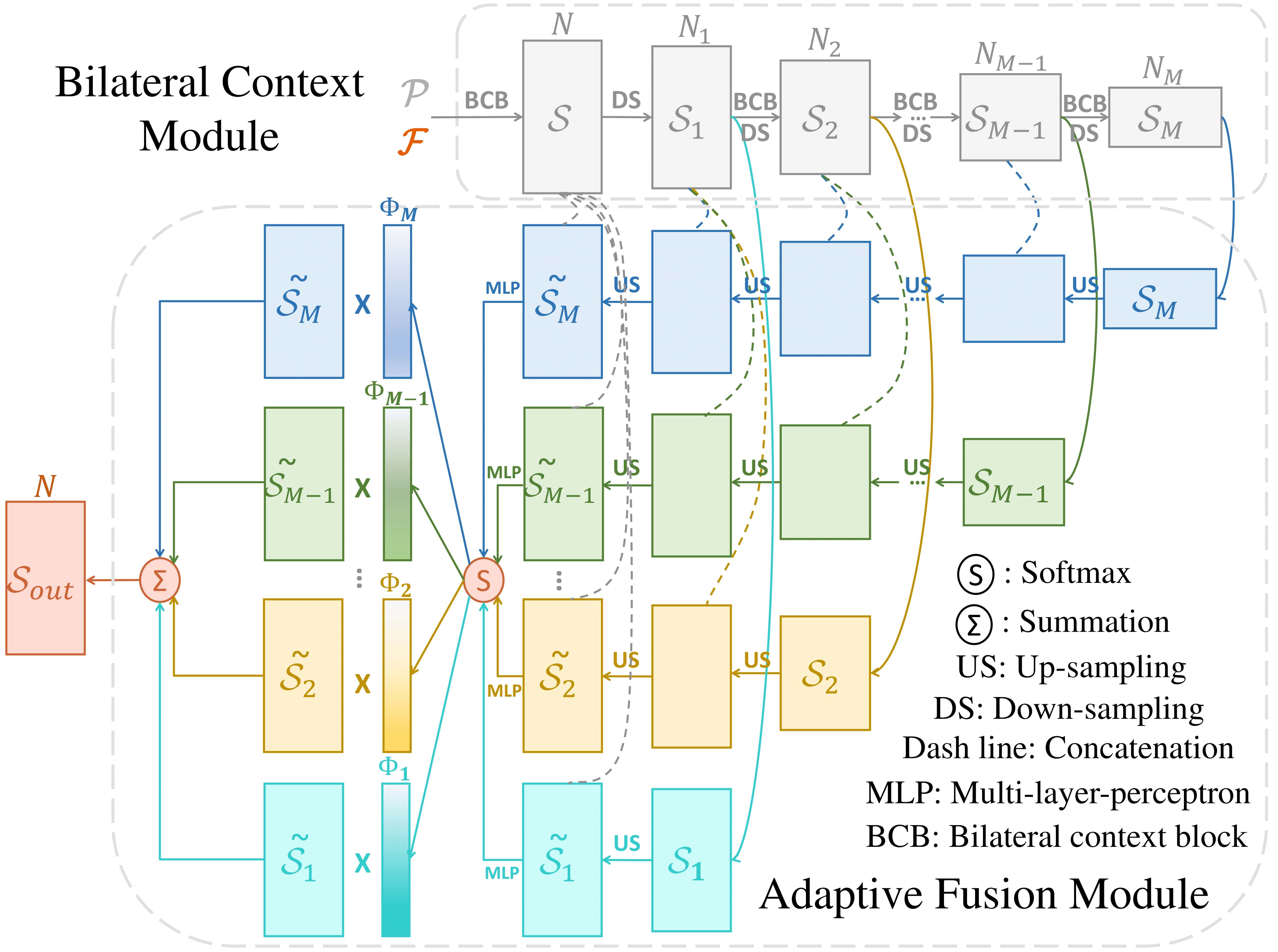}
\end{center}
\caption{The architecture of the \ourdecoder. All the annotations are consistent with the items in Sec.~\ref{sec:metho} of the main paper.}
\label{fig:fusion}
\end{figure}

\begin{table*}
\begin{center}
\resizebox{0.98\textwidth}{!}{
\begin{tabular}{c|c c c| c c c c c c c c c c c c c}
\Xhline{3\arrayrulewidth}\xrowht{15pt}
Testing Area &\textbf{mAcc}&\textbf{OA}&\textbf{mIoU}&ceiling&floor&wall&beam&column&window&door&table&chair&sofa&bookcase&board&clutter\\\hline\xrowht{12pt}
Area 1 &87.7&89.5&76.3&96.5&95.4&80.3&65.4&58.8&78.0&84.3&70.7&82.9&78.0&60.9&73.2&67.9\\\xrowht{12pt}
Area 2 &71.1&86.6&57.8&87.1&95.1&80.0&19.8&33.3&47.5&69.3&45.6&83.1&52.8&50.7&33.1&54.4\\\xrowht{12pt}
Area 3 &89.7&91.7&80.0&95.8&98.2&83.3&74.4&40.5&86.0&88.5&74.4&83.7&79.0&73.6&88.9&73.9\\\xrowht{12pt}
Area 4 &77.9&86.1&64.3&94.8&97.1&78.6&53.0&48.6&30.8&61.0&67.4&77.0&70.1&51.3&44.8&61.6\\\xrowht{12pt}
Area 5 &73.1&88.9&65.4&92.9&97.9&82.3&0.0&23.1&65.5&64.9&78.5&87.5&61.4&70.7&68.7&57.2\\\xrowht{12pt}
Area 6 &92.0&92.5&81.8&96.4&97.5&86.2&79.9&81.0&78.5&90.1&77.1&88.1&65.1&72.4&79.7&71.2\\\hline\xrowht{12pt}
\textbf{6-fold} &83.1&88.9&72.2&93.3&96.8&81.6&61.9&49.5&65.4&73.3&72.0&83.7&67.5&64.3&67.0&62.4\\\Xhline{3\arrayrulewidth}
\end{tabular}
}
\end{center}
\vspace{-3mm}
\caption{Detailed semantic segmentation results (\%) on \emph{S3DIS}~\cite{armeni2017joint} dataset. (\textbf{mAcc}: average class accuracy, \textbf{OA}: overall accuracy, \textbf{mIoU}: mean Intersection-over-Union.\enquote{6-fold}: 6-fold cross-validation result.)}
\label{tab:s3dis_detail}
\end{table*}
\vspace{3mm}
\begin{table*}
\begin{center}
\resizebox{0.93\textwidth}{!}{
\begin{tabular}{c| c c| c c c c c c c c}
\Xhline{3\arrayrulewidth}
\multirow{2}{*}{Method}&\multirow{2}{*}{\textbf{OA}} &\multirow{2}{*}{\textbf{mIoU}}  &man-made&natural&high&low&\multirow{2}{*}{buildings}&hard&scanning&\multirow{2}{*}{cars}  \\
& &  &terrain&terrain&vegetation&vegetation&&scape&artefacts&   \\\hline\xrowht{10pt}
SnapNet~\cite{boulch2018snapnet} &88.6&59.1&82.0&77.3&79.7&22.9&91.1&18.4&37.3&64.4\\\xrowht{10pt}
ShellNet~\cite{zhang2019shellnet} &93.2&69.3&96.3&90.4&83.9&41.0&94.2&34.7&43.9&70.2\\\xrowht{10pt}
GACNet~\cite{wang2019graph} &91.9&70.8&86.4&77.7&\textbf{88.5}&\textbf{60.6}&94.2&37.3&43.5&77.8\\\xrowht{10pt}
SPG~\cite{landrieu2018large} &94.0&73.2&\textbf{97.4}&92.6&87.9&44.0&83.2&31.0&63.5&76.2\\\xrowht{10pt}
KPConv~\cite{thomas2019kpconv} &92.9&74.6&90.9&82.2&84.2&47.9&94.9&40.0&\textbf{77.3}&\textbf{79.7}\\\xrowht{10pt}
RandLA-Net~\cite{hu2020randla} &\textbf{94.8}&\textbf{77.4}&95.6&91.4&86.6&51.5&\textbf{95.7}&\textbf{51.5}&69.8&76.8\\\hline\xrowht{10pt}
\textbf{Ours}  &94.3&75.3&96.3&\textbf{93.7}&87.7&48.1&94.6&43.8&58.2&79.5   \\ \Xhline{3\arrayrulewidth}
\end{tabular}
}
\end{center}
\vspace{-3mm}
\caption{Semantic segmentation (reduced-8) results (\%) on \emph{Semantic3D}~\cite{hackel2017semantic3d} dataset.}
\label{tab:semantic3d_detail}
\end{table*}
\noindent \textbf{\ourdecoder:}  In addition to Alg.~\ref{alg:fusion} and Sec.~\ref{sec:metho_fuse} in the main paper, we also illustrate the architecture of the \ourdecoder~in Fig.~\ref{fig:fusion} as a complement. As described in Sec.~\ref{sec:impl_decoder} of the main paper, we gradually upsample the extracted feature maps $\{{\mathcal{S}_1}, {\mathcal{S}_2}, {\mathcal{S}_3}, {\mathcal{S}_4}, {\mathcal{S}_5}\}$, respectively. In this case, the upsampled full-sized feature maps are $\{{\tilde{\mathcal{S}_1}}, {\tilde{\mathcal{S}_2}}, {\tilde{\mathcal{S}_3}}, {\tilde{\mathcal{S}_4}}, {\tilde{\mathcal{S}_5}}\}$, all of which are in $\mathbb{R}^{N\times 32}$.

Then, for each upsampled full-sized feature map, we use a fully-connected layer ($\mathrm{FC}$, and its superscript indicates the number of kernels) to summarize the point-level information:
$$
    \phi_m = {\mathrm{FC}}^1({\tilde{\mathcal{S}_m}}), \quad \phi_m\in\mathbb{R}^{N}; 
$$
where ${\forall}{\tilde{\mathcal{S}_m}}\in\{{\tilde{\mathcal{S}_1}}, {\tilde{\mathcal{S}_2}}, {\tilde{\mathcal{S}_3}}, {\tilde{\mathcal{S}_4}}, {\tilde{\mathcal{S}_5}}\}$. Subsequently, we concatenate the $\{\phi_1, \phi_2, \phi_3, \phi_4, \phi_5\}$, and point-wisely normalize them using softmax function:
$$
    \Phi = \mathrm{softmax}\big(\mathrm{concat}(\phi_1, \phi_2, \phi_3, \phi_4, \phi_5)\big), \Phi\in\mathbb{R}^{N\times 5}.
$$
Next, we separate $\Phi$ channel-by-channel, and obtain the fusion parameters: $\{\Phi_1, \Phi_2, \Phi_3, \Phi_4, \Phi_5\}$, all of which are in $\mathbb{R}^{N}$. Hence, the point-level adaptively fused feature map is calculated as:
$$
    \mathcal{S}_{out} = \Phi_1\times {\tilde{\mathcal{S}_1}} + \Phi_2\times {\tilde{\mathcal{S}_2}} + \Phi_3\times {\tilde{\mathcal{S}_3}} + \Phi_4\times {\tilde{\mathcal{S}_4}} + \Phi_5\times {\tilde{\mathcal{S}_5}},  
$$
where $\mathcal{S}_{out}\in\mathbb{R}^{N\times 32}$.

\subsection{Predictions}
Based on $\mathcal{S}_{out}$, we utilize three fully-connected layers and a drop-out layer ($\mathrm{DP}$, and the drop-out ratio shows at the superscript) to predict the confidence scores for all $Q$ candidate semantic classes:
$$
    \mathcal{V}_{pred} = \mathrm{FC}^{Q}\bigg(\mathrm{DP}^{0.5}\Big(\mathrm{FC}^{32}\big(\mathrm{FC}^{64}(\mathcal{S}_{out})\big)\Big)\bigg), 
$$
where $\mathcal{V}_{pred}\in\mathbb{R}^{N\times Q}$.
\subsection{Loss Function} Eq.~\ref{equ:all_loss} of the main paper formulates the overall loss $\mathcal{L}_{all}$ of our network based on the cross-entropy loss $\mathcal{L}_{CE}$ and the augmentation loss $\mathcal{L}_m$ for each \ourblock.

In practice, our \ourencoder~gradually processes a decreasing number of points ($N\rightarrow\frac{N}{4}\rightarrow\frac{N}{16}\rightarrow  \frac{N}{64}\rightarrow\frac{N}{256}$) through five blocks. Empirically, we set the weights $\{0.1, 0.1, 0.3, 0.5, 0.5\}$ for the corresponding five augmentation losses, since we aim to provide more penalties for lower-resolution processing. Therefore, the overall loss for our network is:
$$
    \begin{aligned}
            \mathcal{L}_{all} = &\mathcal{L}_{CE} +\\ &0.1\cdot\mathcal{L}_1 + 0.1\cdot\mathcal{L}_2 +\\ 
            &0.3\cdot\mathcal{L}_3 + 0.5\cdot\mathcal{L}_4 + 0.5\cdot\mathcal{L}_5.
    \end{aligned}
$$

\section{Experiments}
\subsection{Areas of S3DIS}
We include more experimental data about our network's semantic segmentation performance. To be specific, Tab.~\ref{tab:s3dis_detail} shows our results for each area in the S3DIS dataset, including overall accuracy, average class accuracy, and concrete IoUs for 13 semantic classes. To evaluate each area, we apply the rest five areas as the training set.

\subsection{Reduced-8 Semantic3D}
Further, Tab.~\ref{tab:semantic3d_detail} presents our online evaluation results on the smaller test set (\ie, reduced-8, which has four scenes including about 0.1 billion points) of the Semantic3D dataset. Comparing with Tab.~\ref{tab:semantic3d} in the main paper (\ie, results of semantic-8, which contains 15 scenes with 2 billion points), we conclude that our semantic segmentation performance regarding large-scale data is relatively better.

\begin{figure*}
\begin{center}
\includegraphics[width=0.98\textwidth]{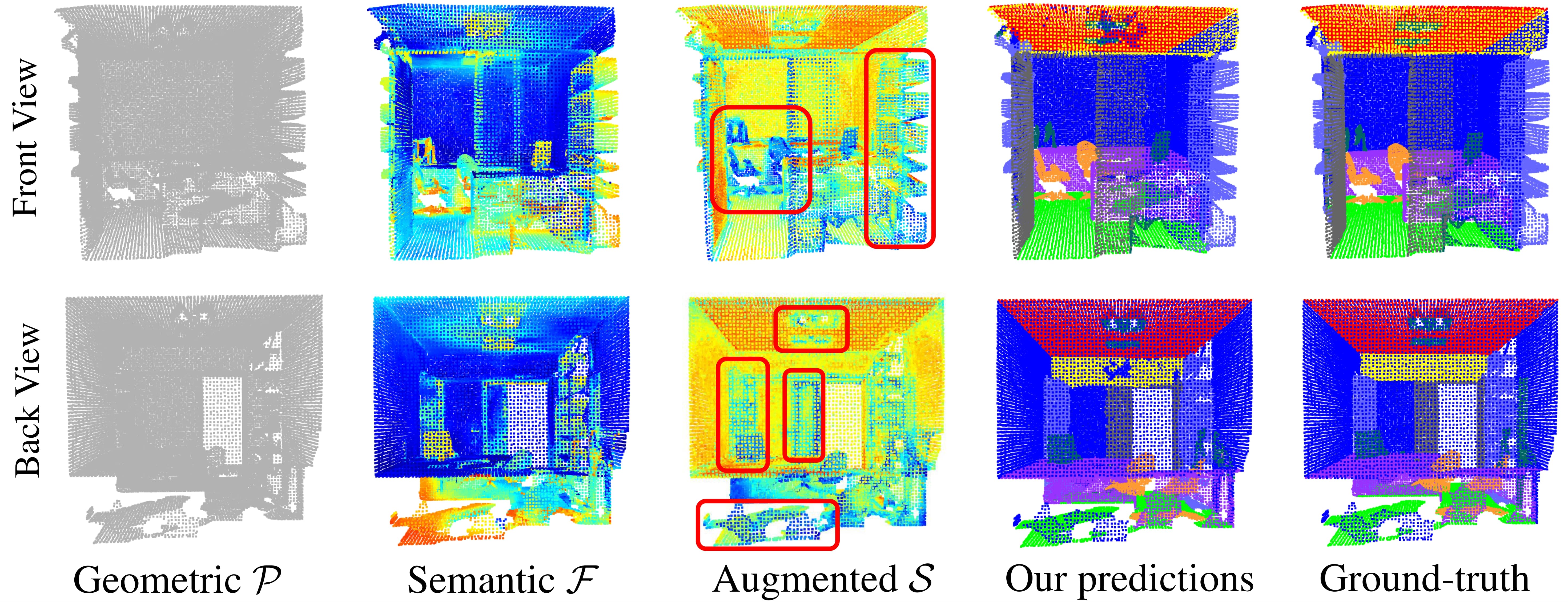}
\end{center}
\caption{Visualization of intermediate features and semantic segmentation results for an office scene in \emph{S3DIS}~\cite{armeni2017joint} dataset. $\mathcal{P}$ denotes the 3D coordinates of the point cloud, and $\mathcal{F}$ presents the semantic information acquired by the Feature Extractor (Sec.~\ref{sec:impl_extractor} in the main paper). Further, $\mathcal{S}$ means the output of our Bilateral Context Block (Sec.~\ref{sec:metho_cross}).}
\label{fig:vis_bilateral}
\end{figure*}

\subsection{Ablation Study}
In addition to the specific ablation studies (Sec.~\ref{sec:exp_abl} in the main paper) about our \ourblock~and \ourdecoder~respectively, we also conduct an ablation study to investigate some variants of our network:
\begin{table}
\begin{center}
\resizebox{0.92\columnwidth}{!}{
\begin{tabular}{c| l c}
\Xhline{3\arrayrulewidth}\xrowht{10pt}
Model &Description &mIoU (\%)\\\hline\xrowht{6pt}
$\mathrm{N}_0$ &Baseline model&60.8\\\xrowht{6pt}
$\mathrm{N}_1$ &Efficient model&64.8\\\xrowht{6pt}
$\mathrm{N}_2$ &Dilated model&62.5\\\xrowht{6pt}
$\mathrm{N}_3$ &Equal-weighted model&64.0\\\xrowht{6pt}
$\mathrm{N}_4$ &Simplified model&63.5\\\hline\xrowht{6pt}
$\mathbf{\mathrm{N}_5}$ &\textbf{Proposed model}  &\textbf{65.4}\\ \Xhline{3\arrayrulewidth}
\end{tabular}
}
\end{center}
\vspace{-2mm}
\caption{Ablation study about different variants of our network, tested on Area 5, \emph{S3DIS}~\cite{armeni2017joint} dataset.}
\label{tab:supp_abl}
\end{table}
\begin{itemize}
 \item \textbf{Baseline model:} We replace both our \ourblock~and \ourdecoder~with their baseline forms, which are explained in the ablation studies of the main paper.
 \item \textbf{Efficient model:} We apply the random sampling instead of the Farthest Point Sampling (FPS).
 \item \textbf{Dilated model:} We use dilated-knn~\cite{engelmann2020dilated} to search the neighbors of each point, in order to increase the size of point's receptive field. The dilated factor $d=2$. 
 \item \textbf{Equal-weighted model:} We set an equal weight ($\omega_i = 0.3$) for all of the augmentation losses in Eq.~\ref{equ:all_loss} (\ie, calculating the overall loss $\mathcal{L}_{all}$) of the main paper.
 \item \textbf{Simplified model:} We only study four resolutions of the point cloud through the \ourencoder. The number of points decreases as: $N\rightarrow \frac{N}{4}\rightarrow \frac{N}{16}\rightarrow \frac{N}{64}$, while the number of channels goes as: $16\rightarrow 64\rightarrow 128\rightarrow 256$. 
\end{itemize}

Tab.~\ref{tab:supp_abl} indicates that such an efficient random sampling ($\mathrm{N}_1$) cannot perform as effectively as FPS does since the randomly sampled subsets can hardly retain the integrity of inherent geometry. As there is always a trade-off between the network's efficiency and effectiveness, we look forward to better balancing them in future work. Besides, increasing the size of the point's receptive field ($\mathrm{N}_2$) as~\cite{engelmann2020dilated} may not help in our case. Further, we observe that it is not optimal to use the equal-weighted \ourblocks~ ($\mathrm{N}_3$) for multi-resolution point clouds. Moreover, our network can be flexibly assembled: for an instance of model $\mathrm{N}_4$ that consists of fewer blocks, even though the performance is reduced, it consumes less GPU memory. 

\begin{table}
\begin{center}
\resizebox{\columnwidth}{!}{
\begin{tabular}{|l|c|c|c|c|c|c|}
\hline
\multicolumn{2}{|l|}{Layer} & 1 & 2 & 3 & 4 & 5 \\ \hline
\multicolumn{2}{|l|}{\#Points}      &40960   &10240   &2560   &640   &160   \\ \hline
\multirow{2}{*}{3D Space}     &\textbf{Mean}    &$\downarrow12$   &$\downarrow24$   &$\downarrow47$   &$\downarrow85$   &$\downarrow154$   \\ \cline{2-7} 
                      & \textbf{Variance}     &$\downarrow0.1$   &$\downarrow0.2$   &$\downarrow0.5$   &$\downarrow2$   &$\downarrow13$   \\ \hline
\multirow{2}{*}{Feature Space}     &\textbf{Mean}    &$\downarrow45$   &$\downarrow693$   &$\downarrow814$   &$\downarrow124$   &$\downarrow317$   \\ \cline{2-7} 
                      &\textbf{Variance}    &$\downarrow11.9$   &$\downarrow16.3$   &$\downarrow24.7$   &$\downarrow46.2$   &$\downarrow104$  \\ \hline
\end{tabular}
}
\end{center}
\vspace{-2mm}
\caption{The general \emph{changes} ($\times10^{-3}$) of neighborhoods by involving bilateral offsets.}
\label{tab:bilateral_behave}
\end{table}

\section{Visualization}
\subsection{Bilateral Context Block}
In Fig.~\ref{fig:vis_bilateral}, we present the \ourblock's output features in a heat-map view. Particularly, we observe that the Bilateral Context Block can clearly raise different responses for close points (in red frames) that are in different semantic classes.

Besides, we calculate the average neighbor-to-centroid Euclidean-distances and average neighborhood variances in 3D space (Eq.~\ref{eq:shift} in the main paper) and feature space (Eq.~\ref{eq:shift_feat}), using the S3DIS samples. Tab.~\ref{tab:bilateral_behave} shows that shifted neighbors get closer to centroids as expected, in both 3D and feature spaces. Further, the variances inside the neighborhoods also drop. In general, the shifted neighbors tend to form compact neighborhoods.

\subsection{Visualizations and Failure Cases}
\begin{figure*}
\begin{center}
\includegraphics[width=0.93\textwidth]{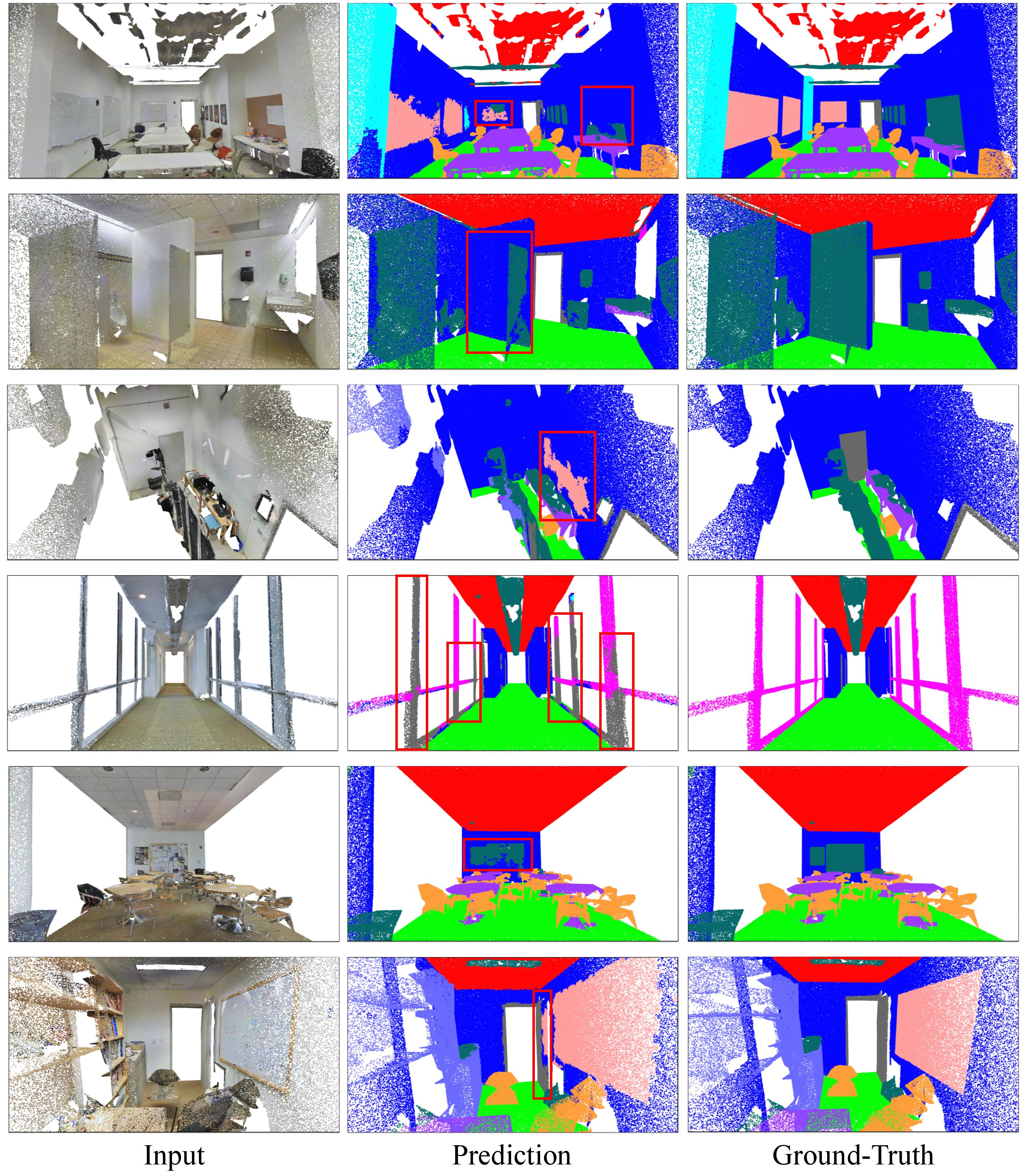}
\end{center}
\caption{Examples of our semantic segmentation results of \emph{S3DIS}~\cite{armeni2017joint} dataset. The first column presents the input point cloud scenes (\enquote{Input}) of some indoor rooms. The second column shows the semantic segmentation predictions of our network (\enquote{Prediction}), while the last column indicates the ground-truths (\enquote{Ground-Truth}). The main differences are highlighted in red frames.}
\label{fig:vis_s3dis}
\end{figure*}

We provide more visualizations of our semantic segmentation network's outputs and some failure cases. Specifically, Fig.~\ref{fig:vis_s3dis} presents our results on six different types of rooms, which are \emph{conference, WC, storage, hallway, lobby, office} rooms, respectively. Unfortunately, we find that the proposed method is not competent enough for distinguishing the objects that are in similar shapes. The main reason is that the network relies on the local neighborhood of each point, while lacks the geometric information about the specific object that each point belongs to. In the 3rd row of Fig.~\ref{fig:vis_s3dis}, \emph{beam} is incorrectly classified as \emph{door} since it looks like the doorframes; while \emph{wall} is wrongly predicted as \emph{board} or \emph{clutter} in the rest of rows.  

\begin{figure*}
\begin{center}
\includegraphics[width=0.9\textwidth]{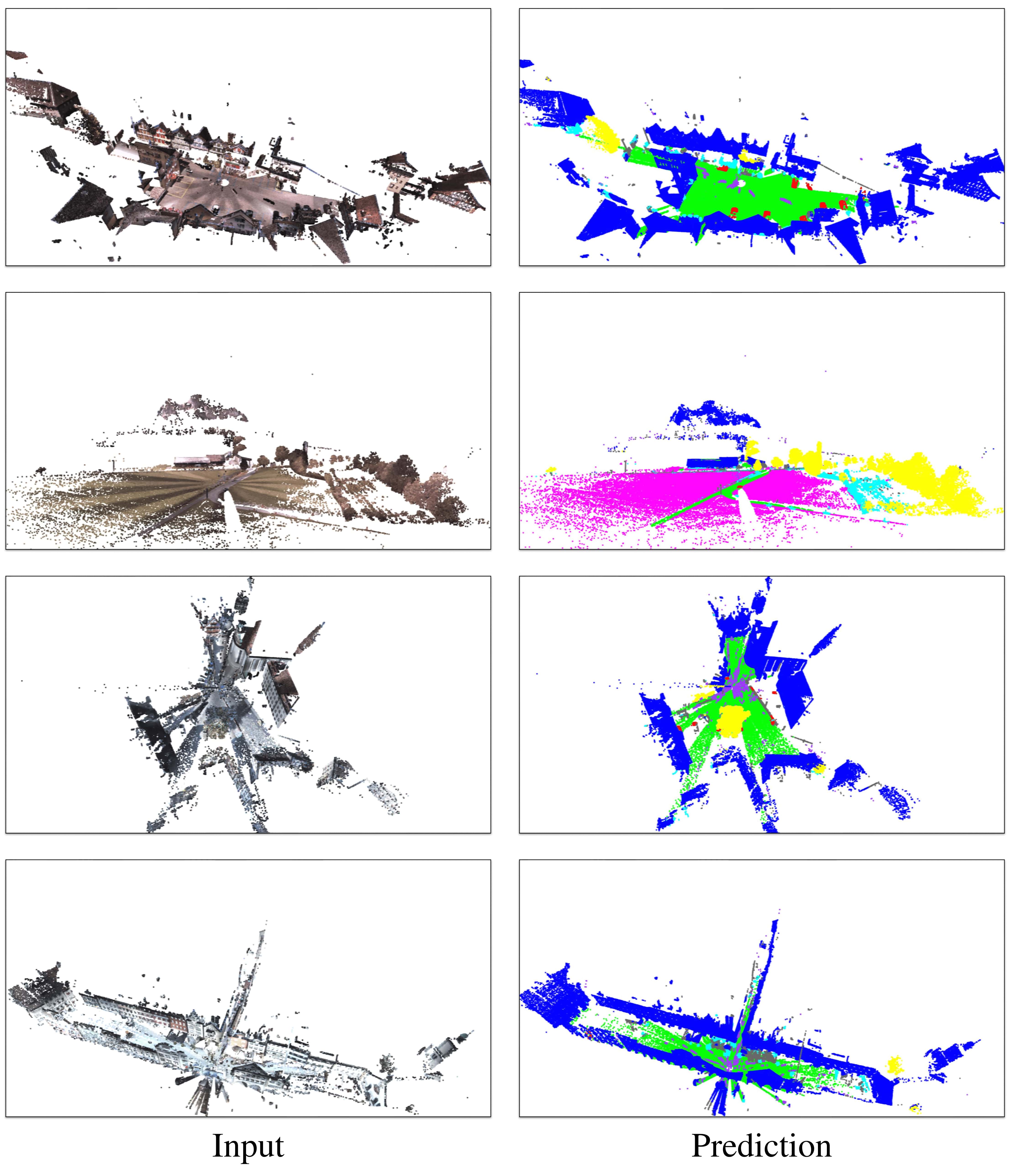}
\end{center}
\caption{Examples of our semantic segmentation predictions of \emph{Semantic3D}~\cite{hackel2017semantic3d} dataset. The first row is about an urban square, the second one shows a rural farm, the third one illustrates a cathedral scene, and the last is scanned from a street view.}
\label{fig:vis_semantic3d}
\end{figure*}

In Fig.~\ref{fig:vis_semantic3d}, we show the general semantic segmentation performances on some large-scale point clouds of typical urban and rural scenes. Although the ground-truths of Semantic3D's test set are unavailable, our semantic predictions of these scenes are visually plausible.

\begin{figure*}
\begin{center}
\includegraphics[width=\textwidth]{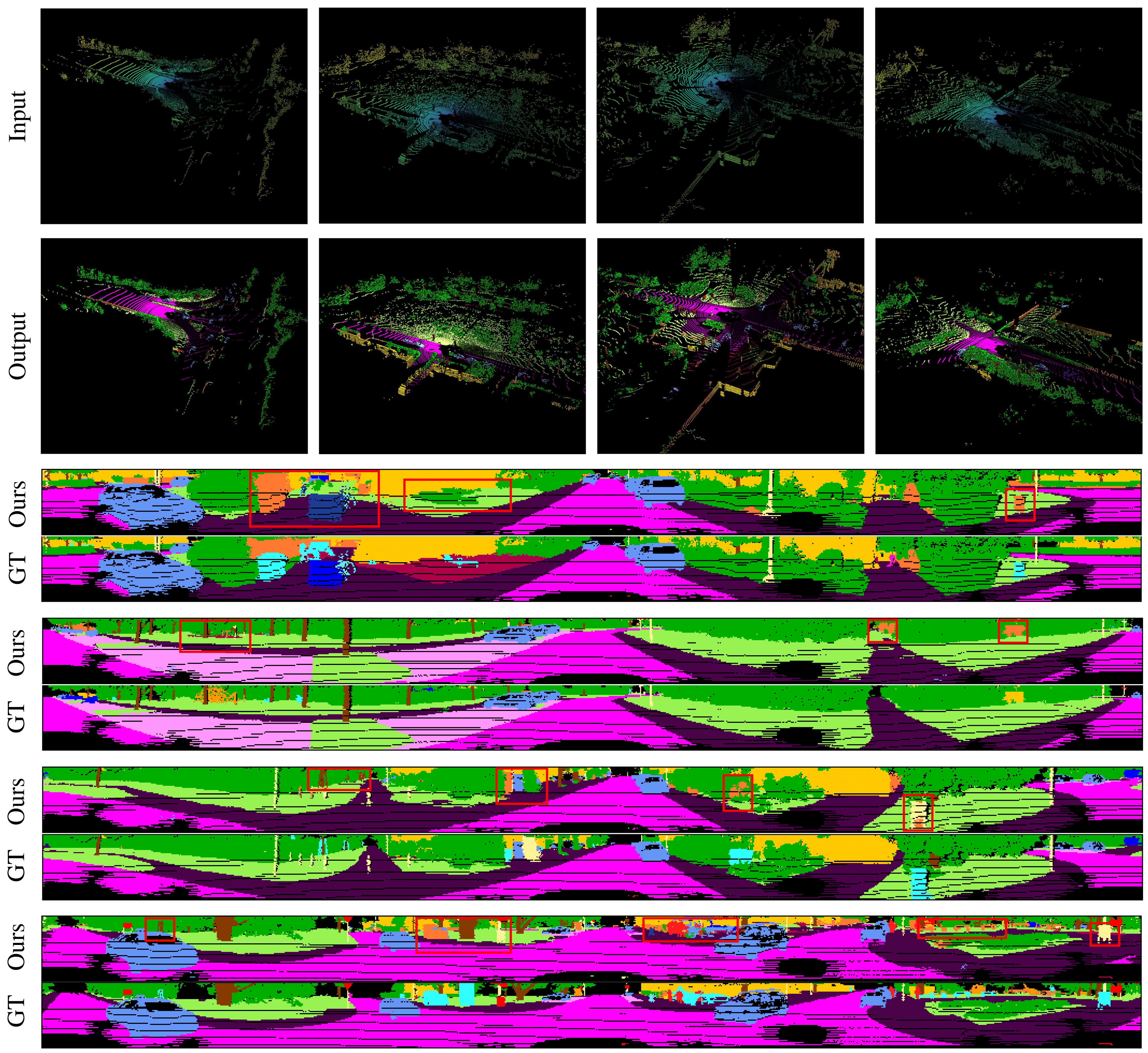}
\end{center}
\caption{Examples of our semantic segmentation predictions of \emph{SemanticKITTI}~\cite{behley2019semantickitti} dataset. The first two rows show the general 3D views of the input traffic scenarios (\enquote{Input}) and our semantic segmentation outputs (\enquote{Output}), respectively. The remaining rows compare our predictions (\enquote{Ours}) and the ground-truths (\enquote{GT}) in 2D panorama views, where the failure cases are highlighted in red frames.}
\label{fig:vis_kitti}
\end{figure*}

In addition, we compare our results against the ground-truths on the validation set (\ie, Sequence 08) of SemanticKITTI dataset in Fig.~\ref{fig:vis_kitti}. Particularly, we illustrate some 3D point cloud scenes in the views of 2D panorama, in order to clearly show the failure cases (highlighted in red color). In fact, the proposed network is able to find some small objects that are semantically different from the background, however, the predictions are not accurate enough since we only use the 3D coordinates as input. As SemanticKITTI is made up of the sequences of scans, in the future, we will take the temporal information into account.

\clearpage

\clearpage

\clearpage

\end{document}